\documentclass[10pt,twocolumn,letterpaper]{article}

\usepackage{cvpr}              

\usepackage{xspace}
\newcommand{\method}{IR4RL\xspace}

\usepackage[ruled]{algorithm2e} 
\SetEndCharOfAlgoLine{}
\SetCommentSty{mycommfont}
\SetKwComment{tcc}{\hfill$\triangleright$~}{}  
\SetKwComment{tcp}{\hfill$\triangleright$~}{}  

\SetKwInput{KwInput}{Inputs}
\SetKwInput{KwHyper}{Hyperparameters}
\SetKwInput{KwAlgo}{Algorithm}
\SetKwInput{KwOutput}{Output}

\SetAlFnt{\small}
\SetAlCapFnt{\small}
\SetAlCapNameFnt{\small}
\SetAlCapHSkip{0pt}
\usepackage{float}     
\usepackage{bbm}
\usepackage{makecell}
\usepackage{lineno}

\definecolor{cvprblue}{rgb}{0.21,0.49,0.74}
\usepackage[pagebackref,breaklinks,colorlinks,allcolors=cvprblue]{hyperref}

\def\confName{CVPR}
\def\confYear{2027}

\title{IR4RL: Reinforcement Learning from Intermediate Renders for Image-to-Code Generation\vspace{-1em}}

\author{
Omri Kaduri$^1$\thanks{Equal contribution, order determined at random.} \quad 
Kate Feingold$^{1}$\footnotemark[1] \quad Phillip Isola$^2$ \quad Tali Dekel$^1$ \\
$^1$ Weizmann Institute of Science \quad $^2$MIT\\
{\fontsize{10.5}{12}\selectfont Project page: \href{https://ir4rl.github.io}{ir4rl.github.io}}
\vspace{-1em}
}

\begin{document}

\twocolumn[{%
  \renewcommand\twocolumn[1][]{#1}%
  \maketitle

  \begin{center}
  \vspace{-20pt}
    \includegraphics[width=0.9\textwidth]{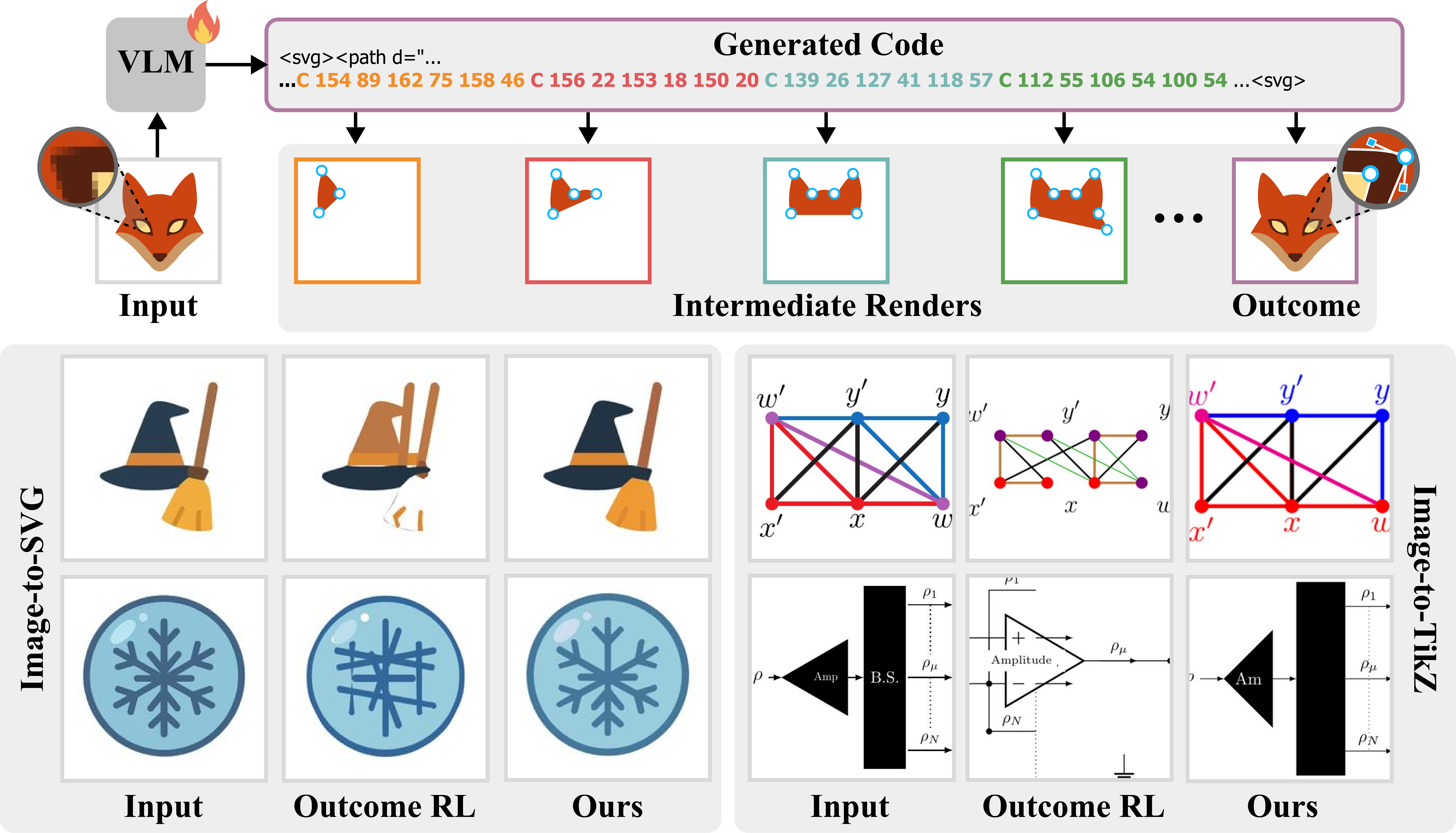}

    \captionof{figure}{We introduce a method that leverages
    \textbf{I}ntermediate \textbf{R}enders for
    \textbf{R}einforcement \textbf{L}earning (\textbf{IR4RL})
    post-training of image-to-code VLMs as a supervision signal,
    allowing us to improve over outcome-only RL and achieve
    state-of-the-art results in Image-to-SVG and Image-to-TikZ tasks.}
    \label{fig:teaser}
  \end{center}
}]

\begingroup
\renewcommand\thefootnote{}%
\footnotetext{$^*$Equal contribution, order determined at random.}%
\endgroup

\begin{abstract}
Reinforcement learning is increasingly used to post-train vision-language models for image-to-code generation, such as generating SVG code from a reference image, by optimizing rewards computed from the final rendered output. However, relying on a single terminal reward provides sparse feedback that is poorly aligned with the contribution of individual tokens. A generated program may contain operations that accurately reproduce some parts of the target image alongside others that introduce errors, yet all tokens are trained from the same final outcome.  We observe that many intermediate code prefixes are not only executable, but already produce meaningful partial renders that reflect progress toward the target. This property provides a natural source of denser supervision during generation. Based on this observation, we introduce \textbf{IR4RL}, an RL framework with a token-level render-progress reward that turns changes between intermediate renders into localized feedback for the generated sequence. We evaluate our approach on Image-to-SVG and Image-to-TikZ generation. Across both tasks, our method improves over supervised fine-tuning and standard GRPO, yielding new state-of-the-art open-source models. This shows that intermediate rendering provides a simple and effective source of process supervision for RL post-training of image-to-code models. 

\end{abstract}


\vspace{-15pt}
\section{Introduction}

Recent generative image models synthesize illustrations, icons, and slides directly in pixel space with striking visual quality
\citep{rombach2022high, esser2024scaling}. Yet raster outputs cannot be edited or scaled arbitrarily without loss of quality. Many downstream applications therefore require the underlying \emph{code} elements. Image-to-code generation produces a program whose rendering reproduces the image while making it compatible with existing graphics and design pipelines. Recent vision-language models (VLMs) have made substantial progress on this task, spanning a growing range of representations, including vector graphics
\citep{yang2025omnisvg}, scientific figures
\citep{belouadi2024detikzify, zhao2025vincicoder}, vector animations
\citep{yang2026omnilottie, Chen_2026_CVPR_lottiegpt}, and parametric CAD \citep{chen2025cadcrafter}.

These models are typically first trained with supervised fine-tuning (SFT) on paired image and code data. SFT learns from reference programs, without observing how an error in the code translates into a visual error in the rendered image.  
Rendering-based reinforcement learning has therefore emerged as a natural post-training step, grounding generated code in its visual outcome \citep{rodriguez2026rendering_rlrf, zhao2025vincicoder}. 
However, a program may contain both helpful and harmful decisions, yet a single outcome reward cannot distinguish their individual contributions.

Our key observation is that image-to-code generation exposes meaningful visual feedback during generation. Intermediate prefixes can be rendered as partial programs, revealing how the reconstruction evolves toward the target, as shown in Fig.~\ref{fig:teaser}.
Based on this insight, we introduce \textbf{\method} (\textbf{I}ntermediate \textbf{R}enders for \textbf{R}einforcement \textbf{L}earning), a framework with a render-progress reward that uses the change in visual similarity between consecutive renders as a local signal of the contribution of newly generated code.
We propagate this reward backward to the tokens that produced it and combine the resulting token-level reward with the outcome reward from the final render.

Our approach applies to image-to-code representations where intermediate program states can be rendered and evaluated. We study two such settings, Image-to-SVG and Image-to-TikZ generation, which differ in representation syntax and rendering pipeline.

Across both tasks, our method consistently improves over supervised
fine-tuning and outcome-only RL with GRPO~\citep{deepseek-math}, establishing new state-of-the-art results among open-source models while producing substantially shorter programs. Our analysis shows that process and outcome feedback are complementary, with their combination performing best on both tasks. We further find that finer-grained intermediate feedback improves performance and that our method shifts probability toward high-quality generations, making them easier to obtain with limited test-time scaling. These results highlight intermediate renderings as an effective
source of process-level supervision for image-to-code post-training.

In summary, our contributions are: 
\begin{itemize} 
    \item \textbf{Render-progress reward.} We identify intermediate renderings as a source of process-level supervision for image-to-code generation, and introduce a reward that turns changes between consecutive renders into render-progress rewards.
    \item \textbf{Efficient RL method for image-to-code post-training.} Building on the render-progress reward, we propose an RL method \method and demonstrate that it outperforms standard outcome-based supervision, successfully generalizing across two image-to-code tasks. We study its behavior across design choices, including combining with outcome reward and supervision granularity. 
    \item \textbf{Strong results on Image-to-SVG and Image-to-TikZ.} Extensive evaluations demonstrate consistent gains on Image-to-SVG and Image-to-TikZ tasks over baselines, establishing state-of-the-art results among open-source models while producing substantially shorter programs.
\end{itemize}
The model weights for both tasks will be open-sourced upon publication. 

\section{Related work}
\label{sec:related_work}
\paragraph{{\bf Image-to-code tasks.}}
Image-to-code generation converts an image into a program whose rendering reproduces the input, spanning vector graphics, scientific figures, webpages, and CAD~\citep{yang2025omnisvg,belouadi2024detikzify,yun2024web2code,chen2025img2cad,zhao2026beyondnl2code}.
Image-to-SVG supports scalable and editable graphics.
Optimization-based methods such as DiffVG~\citep{DiffVG} and LIVE~\citep{xu2022live} directly optimize rendered similarity and achieve strong reconstruction, but require slow per-image optimization.
VLM-based approaches such as StarVector~\citep{rodriguez2023starvector}, OmniSVG~\citep{yang2025omnisvg}, and InternSVG~\citep{wang2025internsvg} instead learn to generate SVG programs from large-scale image-code pairs.
Image-to-TikZ methods similarly generate executable programs for diagrams and scientific figures~\citep{belouadi2024detikzify,saito2025sketch2diagram,zhao2025vincicoder}, alongside related chart- and document-to-code tasks~\citep{tan2025chartmaster,ling2025table2latexrl}. These learned models are trained on reference code with SFT, without visual grounding in the rendered output. 

\vspace{-12pt}
\paragraph{{\bf Learning from rendering feedback.}}
Rendering provides a natural way to supervise image-to-code models beyond matching reference programs. Recent approaches therefore post-train models using rewards computed from generated renderings, including Image-to-SVG~\citep{rodriguez2026rendering_rlrf,wang2026ctrls} and Image-to-TikZ~\citep{zhao2025vincicoder,zeng2026davinci}.
However, these rewards are aggregated at the sequence level, even when augmented with signals for format, code efficiency, language alignment, structural consistency, or compilation success ~\citep{wang2026ctrls,zhao2025vincicoder,zeng2026davinci}, collapsing informative visual progress into a single outcome-level signal. Beyond image-to-code, stroke-based painting agents use per-action canvas improvements as training rewards~\citep{huang2019learning}.
Within image-to-code, intermediate renders have been exploited during inference. 
Iterative methods render partial or completed programs and feed the visual state back to the model for refinement~\citep{liang2026renderintheloop,deng2026visrefiner,yang2026ui2coden}, while search-based approaches use MCTS for Image-to-TikZ~\citep{belouadi2024detikzify} or ERM for vision-to-code~\citep{liu2026visualerm}.
Our approach instead uses intermediate renders during \emph{training}, propagating visual-progress feedback from renderable code prefixes to their generating tokens, without changing inference.

\paragraph{{\bf Fine-grained credit assignment.}}
Fine-grained credit assignment has been extensively studied in mathematical reasoning, where outcome reward models score solutions while process reward models provide feedback at intermediate reasoning steps~\citep{cobbe2021training,uesato2022solving}. Such supervision has been obtained from human step-level labels~\citep{lightman2024verify}, automatic step-wise supervision~\citep{wang2024mathshepherd}, and learned estimates of the process~\citep{setlur2025rewarding}. In theorem proving, Lean has similarly been used directly as a process oracle, converting tactic-level verification into fine-grained RL feedback without a learned reward model~\citep{kim2026process}. Related ideas also appear beyond mathematical reasoning: code-generation methods use compiler and execution feedback for finer-grained optimization~\citep{dou2024stepcoder,ye2025process}, while RLHF-V uses segment-level human corrections for dense multimodal alignment~\citep{yu2024rlhf}.
Potential-based reward shaping~\citep{ng1999policy} modifies a reward of a general MDP with a difference of state potentials under conditions that preserve policy optimality. GAE~\citep{schulman2015high} further adds variance-reducing discounting to this model at the cost of tolerable bias in gradient estimation. 

A central challenge is obtaining reliable intermediate supervision, which may require human annotations, learned or sampling-based verifiers, or domain-specific execution signals. Our key observation is that image-to-code generation exposes directly evaluable intermediate states: many partial programs can already be rendered and compared with the target. We exploit this property to measure changes in visual alignment between successive renders, obtaining localized process rewards without human annotations or a learned verifier.

\section{Preliminary}
\label{sec:prelim}
Given a reference image $x$, a vision-language model $\pi_\theta$ generates a graphics program $y$ whose rendering aims to reconstruct the input. Supervised fine-tuning trains the model to match a reference code $y^*$, potentially penalizing visually accurate programs that differ from it. In contrast, rendering-based reinforcement learning~\citep{zhao2025vincicoder,rodriguez2026rendering_rlrf} does not use ground truth code and rewards visual similarity $S$ between the target and the program's rendering $\mathcal{R}(y)$ with an outcome reward
$R(y,x)=S\bigl(\mathcal{R}(y),x\bigr)$.

\paragraph{Group Relative Policy Optimization.}
For each image prompt $x$, GRPO~\citep{deepseek-math} samples a group of $G$ programs
$\{y_i\}_{i=1}^{G}$ from the current policy $\pi_\theta$ and scores them with
outcome rewards $R_i=R(y_i,x)$. The group-relative outcome advantage is:
\begin{equation}
A_i^{\mathrm{Outcome}} = \frac{R_i-\mu_R}{\sigma_R+\varepsilon},
\end{equation}
where $\mu_R$ and $\sigma_R$ are the group mean and standard deviation, and
$\varepsilon>0$ ensures numerical stability.  With on-policy updates, PPO-style clipping is not used, and
the policy is trained with
\begin{equation}
\begin{aligned}
\mathcal{L}(\theta)
={}&\scalebox{0.88}{$\displaystyle
-\mathbb{E}\Bigl[
\tfrac{1}{G}\sum_{i=1}^{G}
\tfrac{1}{T_i}\sum_{t=1}^{T_i}
A_i^{\mathrm{Outcome}}
\log\pi_\theta(y_{i,t}\mid x,y_{i,<t})
\Bigr]$}\\[-2pt]
&+\beta\,\mathcal{L}_{\mathrm{KL}}(\theta).
\end{aligned}
\label{eq:grpo}
\end{equation}
where $T_i$ is the length of $y_i$, $y_{i,t}$ is its $t$-th token, and $\mathcal{L}_{\mathrm{KL}}$ 
is a regularizer toward a reference policy. 

\begin{figure*}[t!]
  \centering
  \includegraphics[width=\textwidth]{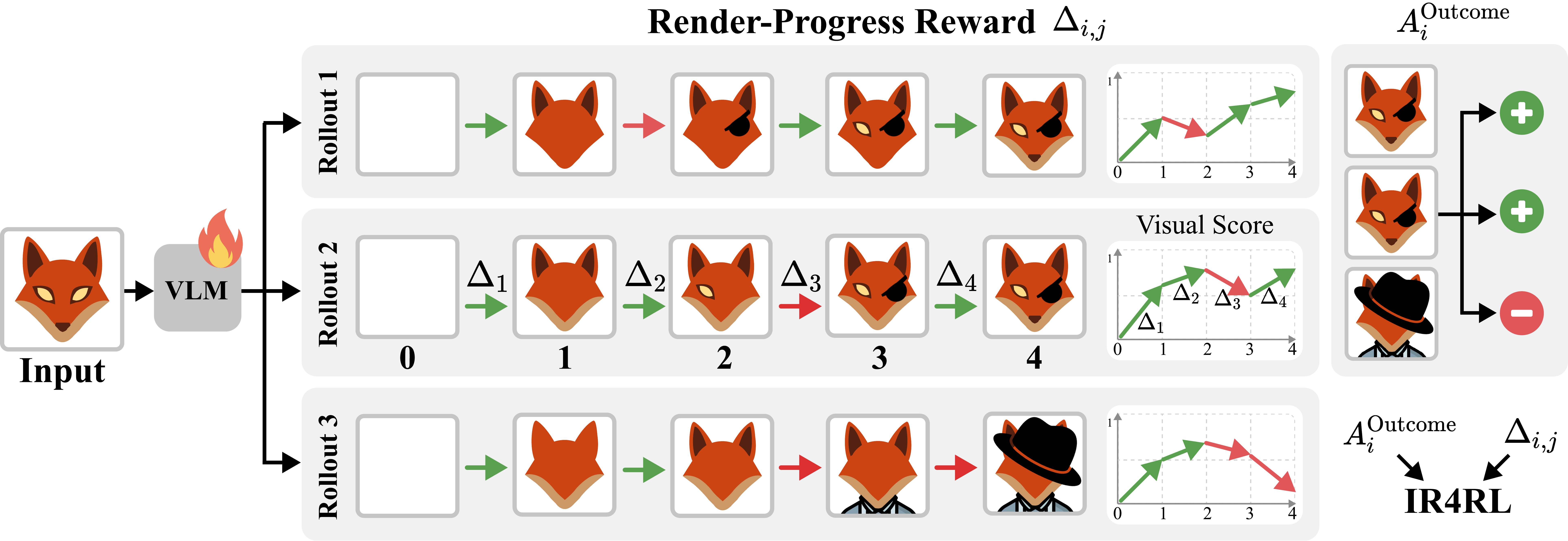}
  \caption{
\textbf{Process Reward from Intermediate Renders.} 
Rollouts 1 and 2 reach similar final quality through different 
generation trajectories, while Rollout 3 makes early progress that later operations reverse. 
Outcome-only advantage rewards all tokens within a rollout equally without distinguishing helpful from harmful decisions.
Our method complements it with localized feedback leveraging changes in visual score between consecutive renders (the arrows $\Delta_j$).
}
\label{fig:intermediate-renders}
\end{figure*}

\section{Method}
\label{sec:method}

The outcome reward evaluates only the final program and therefore cannot localize which decisions helped or hurt the reconstruction. Intermediate renders provide this missing information by revealing how visual quality changes during generation (Fig.~\ref{fig:intermediate-renders}). 

Our method uses these intermediate changes as process supervision, while retaining outcome supervision on the completed program. We first define a render-progress reward from changes in visual score between successive executable prefixes (Sec.~\ref{sec:progress}). We then propagate these rewards through the generated tokens (Sec.~\ref{sec:propagation}) and combine the resulting process signal with the group-relative outcome advantage (Sec.~\ref{sec:combine}).

\subsection{Render-Progress Reward}
\label{sec:progress}

Let $y=(y_1,\ldots,y_T)$ be a generated program. We obtain intermediate renders at token positions $0=b_0<b_1<\dots<b_M=T$, chosen at boundaries supported by the code representation, such as individual line or curve commands in SVG, or ";"-terminated statements in TikZ. Since these prefixes may contain unclosed elements, we apply a simple closure operator $\mathcal{C}$ that makes them renderable by completing the missing syntax (e.g., adding SVG container tags or closing TikZ scope). This yields a renderable prefix $P_j = \mathcal{C}(y_{1:b_j})$, which we evaluate using the same renderer $\mathcal{R}$ and visual similarity function $S$ as for the final program:
\begin{equation}
F_j = S\bigl(\mathcal{R}(P_j),x\bigr).
\label{eq:vis_score}
\end{equation}

Here, $F_0$ is the visual score of an empty canvas and $F_M=R(y,x)$ corresponds to the completed program. 

We want newly generated code to receive a reward that captures its marginal impact on the reconstruction. The absolute score
$F_j$ does not isolate this contribution, since it conflates the progress
established by earlier segments. Thus, we define the \emph{render-progress reward}
as the change in visual score between consecutive renders:
\begin{equation}
\Delta_j = F_j - F_{j-1},
\label{eq:delta}
\end{equation}
where $\Delta_j>0$ rewards a segment that improves alignment with the target,
while $\Delta_j<0$ penalizes one that degrades it (green and red arrows in
Fig.~\ref{fig:intermediate-renders}). 
We analyze the effect of rendering granularity in Fig.~\ref{fig:analysis}c.

\subsection{Propagating Render-Progress Rewards}
\label{sec:propagation}

The delta reward $\Delta_j$ is defined per \emph{segment}, whereas the policy is updated at every \emph{token}.
Tokens within a segment therefore receive no direct reward, even though they affect the render at the next boundary. We address this by propagating future render-progress rewards backward through the sequence with exponentially decaying weights.

For token $t$, let $\mathcal{F}(t)=\{j \mid b_j \ge t\} $ denote the future render events, and let $d(t,j)=b_j-t$ be the token distance from $t$ to event $j$. We aggregate the corresponding render-progress rewards as
\begin{equation} A_t^{\mathrm{Process}} = \sum_{j\in\mathcal{F}(t)} \lambda^{d(t,j)}\Delta_j, \qquad \lambda\in[0,1]. 
\label{eq:progress_credit} 
\end{equation}
The parameter $\lambda$ controls how far each render-progress reward propagates backward. 
We analyze its effect in Fig.~\ref{fig:analysis}b.

\subsection{Combining Process and Outcome Rewards}
\label{sec:combine}

Outcome and process supervision operate at complementary levels. While $A^{\mathrm{Process}}_{i,t}$ provides feedback on changes
in visual quality during generation, $A^{\mathrm{Outcome}}_i$
evaluates the final result relative to other rollouts. A rollout can make positive visual
progress yet end with an incomplete reconstruction, for example
by stopping early. We therefore combine both signals: 
\begin{equation} A_{i,t} = A_i^{\mathrm{Outcome}} + \alpha A_{i,t}^{\mathrm{Process}}, \label{eq:combined_advantage} \end{equation} 
where $\alpha$ controls the contribution of the aggregated render-progress reward. We substitute $A_{i,t}$ for $A_i^{\mathrm{Outcome}}$ in Eq.~\ref{eq:grpo}, so each token is weighted by both final outcome and its visual progress during training.
As shown in Tab.~\ref{tab:quant}, process-only supervision improves reconstruction over outcome-only supervision, while combining both performs best.

In App.~\ref{sec:pbrs_gae}, we interpret our method through PBRS~\citep{ng1999policy} and GAE~\citep{schulman2015high}. 
Viewing render-progress rewards as differences of 
potentials allows us to decompose the shaped return and
its advantage into outcome and process terms. We estimate
them separately using group-relative GRPO and GAE-style temporal weighting respectively.
At $\lambda=1$, the differences telescope,
preserving the optimal policy under the shaped reward, while $\lambda<1$ controls the bias--variance tradeoff of the gradient estimate~\citep{schulman2015high}.

\begin{figure*}[t!]
  \centering
  \includegraphics[width=\textwidth]{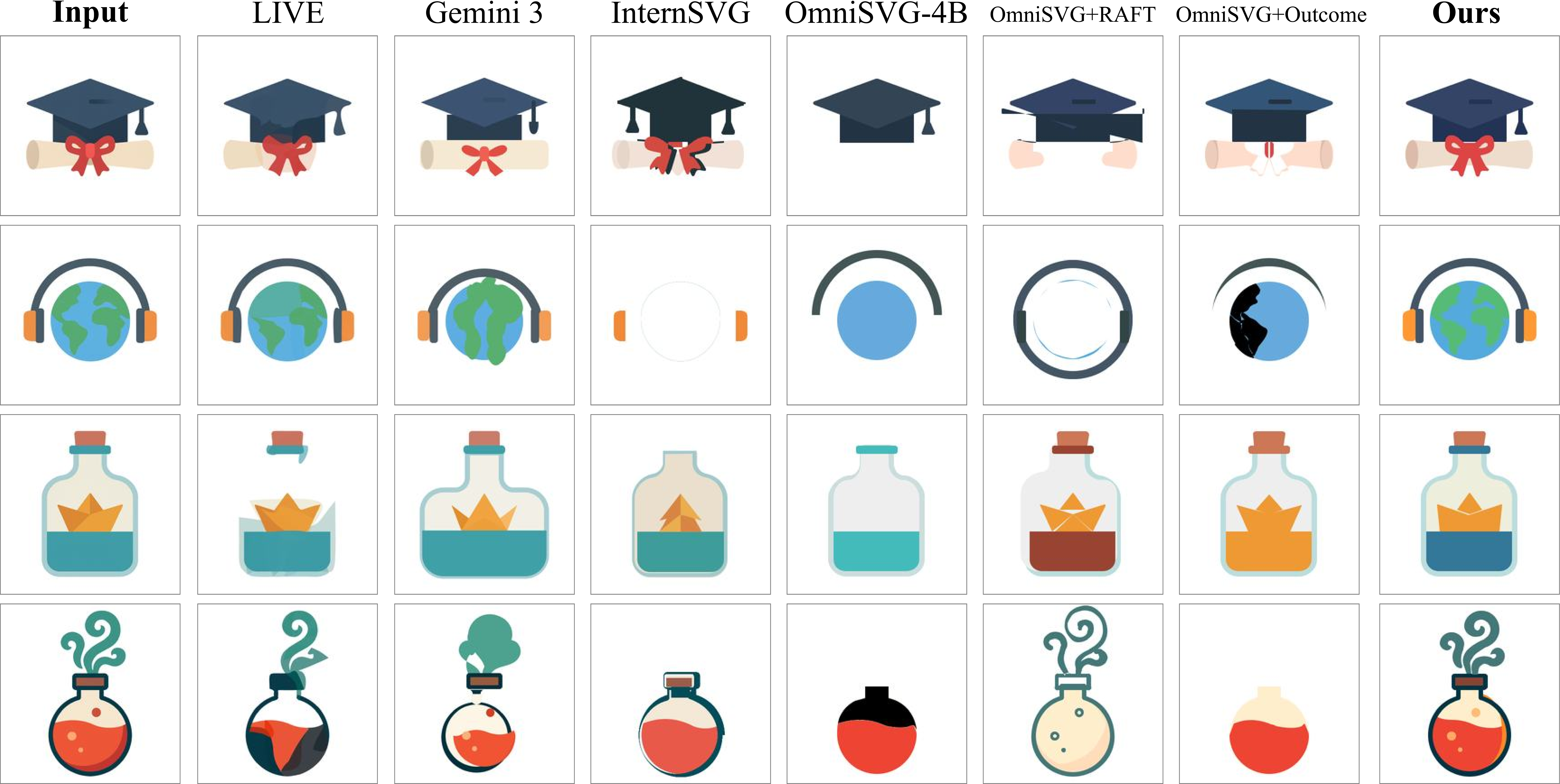}
  \caption{
  \textbf{Qualitative evaluation of Image-to-SVG.} We compare LIVE, Gemini-3-Flash, task-specific SVG models, and post-training baselines. Our method better preserves structure, color, geometry, and fine details, while competing methods more often omit or distort visual elements.
  }
  \label{fig:image_to_svg_qual}
  \includegraphics[width=\textwidth]{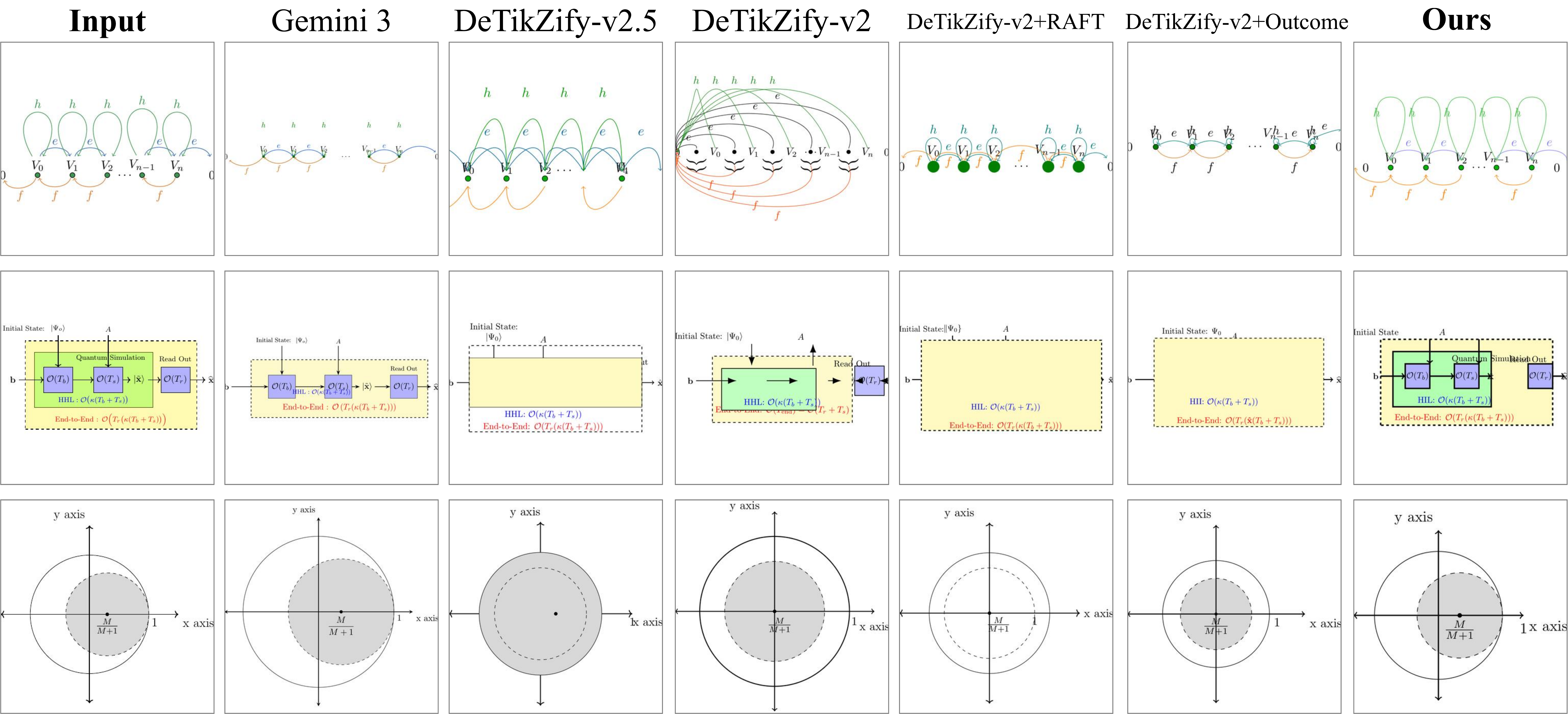}
  \caption{
  \textbf{Qualitative evaluation of Image-to-TikZ.} We compare our method with general-purpose VLMs, task-specific TikZ models, and post-training baselines. Our method better preserves structure, geometry, and annotations, with fewer omitted or distorted elements.
  }
  \label{fig:image_to_tikz_qual}

\end{figure*}
            
\section{Applications}
\label{sec:applications}

We evaluate \method on two image-to-code tasks: Image-to-SVG (Sec.~\ref{sec:svg}) and Image-to-TikZ (Sec.~\ref{sec:tikz}). Both allow intermediate code prefixes to be rendered and
compared with the input image, which is the primary requirement for applying our method. In both representations, sequential commands build up
the image, and later code can cover earlier content through layering
or change its rendered appearance. 
Specifically, our SVG setup, following OmniSVG~\citep{yang2025omnisvg}, represents graphics using paths with line, Bézier, and arc segments, whereas TikZ constructs scientific diagrams by positioning, connecting, and scaling elements.

Across both tasks, we show that intermediate-render supervision consistently improves over supervised fine-tuning and outcome-based post-training. We further analyze the contribution of method components in detail in Sec.~\ref{sec:analysis}.

\subsection{Image-to-SVG}
\label{sec:svg}
\paragraph{Setup.}
We use OmniSVG-4B~\citep{yang2025omnisvg} as our SFT base model. SVG represents vector graphics as sequences of drawing commands and their parameters  
(\emph{e.g.} line \texttt{L62 158}, a Bézier curve \texttt{C ...}), 
making intermediate stages after each such segment renderable (App.~\ref{sec:svg_closure}).
We train on SVG-Stack~\citep{rodriguez2023starvector} train set using the scale-invariant L2 reward of ~\cite{rodriguez2026rendering_rlrf} for a visual score $F$ (Eq.~\ref{eq:vis_score}). All hyperparameters are selected on the SVG-Stack test split, used as our validation set. 
The same parameters also perform
best on our primary test set MMSVGBench~\citep{yang2025omnisvg}
(Fig.~\ref{fig:analysis_extended}a). We report results on MMSVGBench using reconstruction
metrics~\citep{wang2004ssim,zhang2018lpips,oquab2024dinov2,radford2021clip},
aesthetic quality~\citep{wu2023human}, and token length.

We compare against optimization-based methods DiffVG~\citep{DiffVG} and LIVE~\citep{xu2022live}. We also evaluate general-purpose models, including
Qwen3-VL-235B~\citep{Qwen3-VL}, Gemini~3 Flash~\citep{gemini3flash}, Sonnet~5~\citep{claudesonnet5}, and GPT-5.2~\citep{gpt52}, and task-specific
models, including StarVector~\citep{rodriguez2023starvector}, InternSVG~\citep{wang2025internsvg}, and OmniSVG in 4B and 8B sizes. 
Existing outcome-reward (GRPO) baselines \citep{rodriguez2026rendering_rlrf, wang2026ctrls} are not open-source. Hence, to isolate the effect of our reward, we also compare post-training
methods initialized from the same base model. The outcome-only RL baseline uses the final rendering reward, with GRPO as described in Sec.~\ref{sec:prelim}.  RAFT~\citep{dong2023raft} iteratively fine-tunes the model on its highest-scoring samples with a rejection-sampling SFT.
See App.~\ref{sec:setup_svg_sm} for more details on SVG intermediate renderings, training and evaluation.

 \setlength{\tabcolsep}{2pt}
        \begin{table*}[t]
            \centering
            \resizebox{\textwidth}{!}{
            \begin{tabular}{lccccccc|ccccccc}
              \toprule
              & \multicolumn{7}{c}{\textbf{MMSVGBench-Illustrations}} & \multicolumn{7}{c}{\textbf{MMSVGBench-Icons}} \\
              \cmidrule(lr){2-8} \cmidrule(lr){9-15}
                                          & DINO $\uparrow$ & LPIPS $\downarrow$ & MSE $\downarrow$ & SSIM $\uparrow$ & CLIP $\uparrow$ &
      Aesthetic $\uparrow$ & Tokens $\downarrow$ & DINO $\uparrow$ & LPIPS $\downarrow$ & MSE $\downarrow$ & SSIM $\uparrow$ & CLIP $\uparrow$ &
      Aesthetic $\uparrow$ & Tokens $\downarrow$ \\
              \midrule
              \multicolumn{15}{l}{\textit{Optimization-based}} \\
              \midrule
              DiffVG                       & 92.03 & 9.23 & 0.35 & 94.10 & 93.52 & 4.85 & 79.6k & 90.97 & 9.24 & 0.44 & 93.76 & 95.29 & 4.89 & 79.5k \\
              LIVE                         & 94.55 & 10.02 & 0.72 & 95.48 & 93.68 & 4.99 & 8.4k & 94.24 & 9.18 & 0.86 & 95.19 & 95.60 & 4.91 & 8.4k \\
              \midrule
              \multicolumn{15}{l}{\textit{General-purpose VLMs}} \\
              \midrule
              Qwen3-VL-235B                & 92.81 & 28.32 & 5.30 & 87.89 & 89.82 & 4.82 & 5.0k & 92.23 & 29.90 & 7.78 & 84.77 & 91.61 & 4.77 & 5.9k \\
              Gemini 3 Flash               & \underline{96.66} & 18.63 & 2.96 & 90.63 & \underline{95.80} & \textbf{5.09} & 2.6k & 96.92 & 20.14 & 4.39 & 88.58 & \underline{97.47} & \underline{4.97} & 3.5k \\
              Sonnet 5                     & 95.95 & 25.70 & 4.40 & 88.64 & 94.69 & \underline{5.02} & \textbf{0.7k} & 96.58 & 26.01 & 6.09 & 86.61 & 97.10 & 4.94 & \textbf{0.5k} \\
              GPT-5.2                      & 94.71 & 28.97 & 5.30 & 87.41 & 92.90 & \underline{5.02} & \underline{1.7k} & 94.96 & 29.83 & 7.82 & 84.31 & 94.87 & 4.96 & \underline{1.5k} \\
              \midrule
              \multicolumn{15}{l}{\textit{SVG VLMs}} \\
              \midrule
              StarVector-8B                & 85.22 & 25.04 & 5.18 & 89.41 & 83.45 & 4.64 & 2.5k & 88.66 & 25.08 & 6.98 & 86.66 & 89.01 & 4.71 & 2.2k \\
              InternSVG-8B                 & 91.31 & 19.18 & 3.45 & 91.09 & 88.88 & 4.77 & 3.7k & 92.42 & 17.57 & 4.46 & 89.68 & 92.92 & 4.82 & 2.3k \\
              OmniSVG-8B                   & 88.81 & 21.15 & 4.98 & 88.85 & 85.57 & 4.68 & 9.1k & 92.04 & 18.09 & 4.92 & 89.28 & 91.89 & 4.79 & 6.0k \\
              OmniSVG-4B (SFT)             & 85.48 & 22.26 & 5.11 & 89.70 & 82.03 & 4.55 & 11.3k & 89.21 & 19.82 & 5.80 & 87.96 & 88.42 & 4.65 & 8.4k \\
              OmniSVG-4B + RAFT            & 92.20 & 17.56 & 4.60 & 89.32 & 89.34 & 4.81 & 4.8k & 96.55 & 12.21 & 2.57 & 91.55 & 96.05 & 4.93 & 3.3k \\
              OmniSVG-4B + Outcome            & 92.85 & 16.89 & 5.08 & 88.97 & 90.52 & 4.88 & 5.7k & 95.68 & 13.04 & 4.12 & 90.20 & 95.57 & 4.90 & 3.6k \\
              OmniSVG-4B + Process        & 95.89 & \underline{12.28} & \underline{2.27} & \underline{92.68} & 93.57 & 4.99 & 3.5k & \underline{97.35} & \underline{11.27} & \underline{2.20} & \underline{92.11} & 97.03 & 4.95 & 2.8k \\
              OmniSVG-4B + Ours    & \textbf{97.48} & \textbf{9.79} & \textbf{1.17} & \textbf{94.22} & \textbf{96.05} & \textbf{5.09} & 2.5k & \textbf{98.26} & \textbf{8.67} & \textbf{1.23} & \textbf{93.88} & \textbf{98.01} & \textbf{4.98} & 2.1k \\
              \bottomrule
          \end{tabular}
            }
            
            \caption{\textbf{Quantitative evaluation on MMSVGBench.} We compare optimization-based methods, general-purpose VLMs, SVG models, and post-training baselines on the Illustrations and Icons splits. Our method surpasses baselines, substantially improves the SFT base model, and performs best when process and outcome supervision are combined. 
            \textbf{Bold} and \underline{underline} denote the best and second-best results among learned methods, excluding optimization-based methods.
  }
            \label{tab:quant}
            
        \end{table*}

\begin{table}[t]
    \centering
    \setlength{\tabcolsep}{4pt}
    \resizebox{\columnwidth}{!}{
    \begin{tabular}{@{}lccccc@{}}
        \toprule
         & \makecell{Gemini 3\\Flash}
         & \makecell{InternSVG\\8B}
         & \makecell{OmniSVG-4B\\(SFT)}
         & \makecell{OmniSVG-4B\\+ RAFT}
         & \makecell{OmniSVG-4B\\+ Outcome} \\
        \midrule
        Users $\uparrow$ & 60.6\% & 76.3\% & 92.7\% & 74.9\% & 74.0\% \\
        VLM $\uparrow$   & 54.0\% & 74.0\% & 94.0\% & 78.0\% & 76.0\% \\
        \bottomrule
    \end{tabular}
    }
    \caption{\textbf{User study and VLM evaluation.}
    2AFC win rates of our method against each baseline.
    Both humans and a VLM prefer our method
    (VLM--human agreement: 88\%).}
    \label{tab:user_study}
\end{table}

\begin{figure*}[t!]
  \centering
  \includegraphics[width=\textwidth]{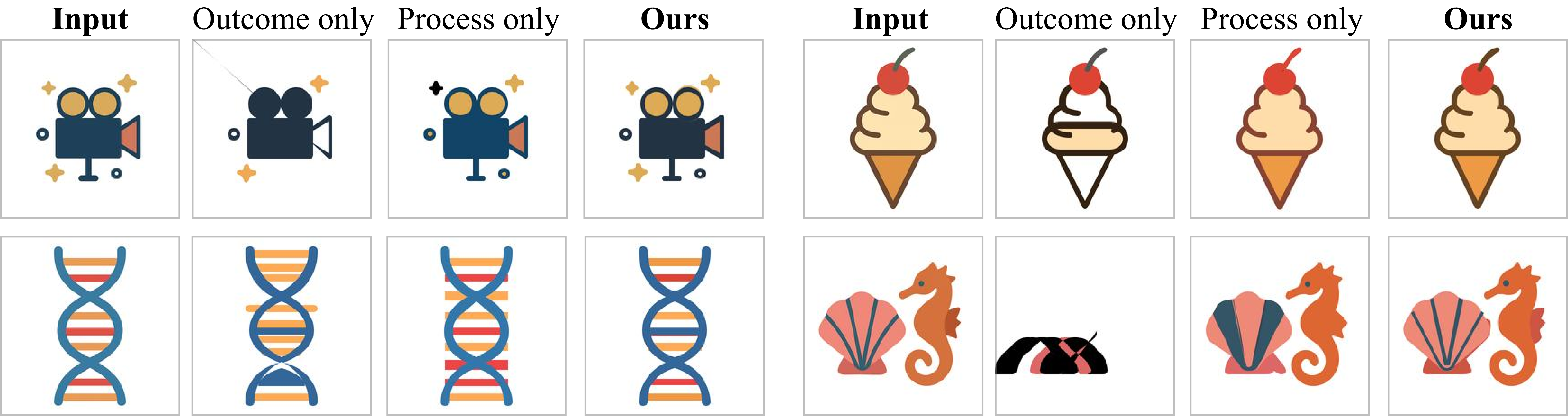}
  \caption{
  \textbf{Qualitative reward composition ablation.} Process-only supervision improves reconstruction over outcome-only supervision, while combining both produces the most faithful reconstructions.
  }
  \label{fig:reward_ablation}
\end{figure*}

\vspace{-10pt}
\paragraph{Results.}
Table~\ref{tab:quant} reports results on MMSVGBench. Our method consistently improves the OmniSVG-4B base model across metrics and outperforms both outcome-only RL and RAFT. 
It also surpasses existing task-specific SVG models, despite having only 4B parameters, and compares favorably with substantially larger general-purpose VLMs. 
Beyond improving reconstruction quality, our model produces considerably shorter SVG programs. 
In addition, in App.~\ref{sec:svg_budget} we show that our gains persist even when outcome RL is given a substantially larger token budget or an explicit code efficiency reward. 

Figure~\ref{fig:image_to_svg_qual} shows representative qualitative comparisons. Our model more faithfully reconstructs the target structure, geometry, and colors, while the baselines more frequently omit or distort visual elements. We complement these metrics with human preferences and VLM-as-a-judge evaluation. As shown in Table~\ref{tab:user_study}, both consistently prefer our method over external and post-training baselines.

\setlength{\tabcolsep}{2pt}
  \begin{table}[t]
      \centering
      \resizebox{\columnwidth}{!}{
      \begin{tabular}{lcccccc}
            \toprule
            & DreamSim $\uparrow$ & SigLIP $\uparrow$ & CLIP $\uparrow$ & LPIPS $\downarrow$ & KID $\downarrow$ & Tokens $\downarrow$ \\
            \midrule
            \multicolumn{7}{l}{\textit{General-purpose (M)LLMs}} \\
            \midrule
            Qwen3-VL-235B                          & 80.5 & 91.7 & 89.2 & 40.3 & 0.75 & 3.9k \\
            Gemini 3 Flash                         & 91.1 & 96.1 & 94.5 & 27.3 & 0.27 & 2.4k \\
            Sonnet 5                               & 85.6 & 94.2 & 92.3 & 35.7 & 0.22 & 0.7k \\
            GPT-5.2                                & 85.1 & 93.9 & 91.6 & 35.5 & 0.44 & 1.5k \\
            \midrule
            \multicolumn{7}{l}{\textit{TikZ VLMs}} \\
            \midrule
            VinciCoder-8B                          & 82.8 & \underline{92.5} & 90.7 & 38.1 & \underline{0.42} & 1.1k \\
            DeTikZify-v2.5                         & 83.7 & 92.3 & \underline{91.3} & 34.5 & 0.58 & \underline{0.8k} \\
            DeTikZify-v2 (SFT)                     & 80.4 & 90.0 & 89.0 & 38.0 & 0.90 & 1.3k \\
            DeTikZify-v2 + RAFT                    & 81.5 & 91.0 & 89.8 & 36.9 & 0.80 & 1.3k \\
            DeTikZify-v2 + Outcome              & \underline{84.2} & 92.4 & \underline{91.3} & \underline{34.0} & 0.45 & 1.0k \\
            DeTikZify-v2 + Ours            & \textbf{86.9} & \textbf{94.1} & \textbf{93.0} & \textbf{32.4} & \textbf{0.30} & \textbf{0.6k} \\
            \bottomrule
        \end{tabular}
      }
      \caption{\textbf{Quantitative evaluation on DaTikZ-v3 test split.}
      Our method improves DeTikZify-v2 and post-training baselines across the visual reconstruction metrics while producing substantially shorter programs. \textbf{Bold} and \underline{underline} mark best and second-best results among task-specific TikZ models.
      }
      \label{tab:quant_tikz}
  \end{table}

\subsection{Image-to-TikZ}
\label{sec:tikz}
\paragraph{Setup.}
For Image-to-TikZ generation, we use DeTikZify-v2~\citep{tikzero2025} as the supervised base model. TikZ programs similarly construct figures through sequences of ";"-separated drawing commands (\emph{e.g.} \verb|\node at (0,2){4};|), allowing intermediate states to be rendered and evaluated after each one (App.~\ref{sec:tikz_closure}). We train on DaTikZ-v3~\citep{tikzero2025} training set and use the DeTikZify's~\citep{belouadi2024detikzify} \emph{SelfSim} reward for a visual score $F$ (Eq.~\ref{eq:vis_score}), following its reported correlation with human judgment of scientific diagrams. 
All hyperparameters are selected on a held-out validation set of 50 samples. 
We evaluate on the DaTikZ-v3 test set using visual reconstruction metrics \citep{zhai2023sigmoid, fu2023dreamsim} and token length, comparing to general-purpose models,
task-specific models, including VinciCoder~\citep{zhao2025vincicoder} and DeTikZify-v2.5. Notably, we surpass VinciCoder-8B and DeTikZify-v2.5, which both represent open-source GRPO baselines. As in
the Image-to-SVG application, we additionally compare against RAFT and
outcome-only RL initialized from the same SFT model.
See App.~\ref{sec:setup_tikz_sm} for more details on TikZ intermediate rendering, training and evaluation parameters. 

\vspace{-10pt}
\paragraph{Results.}
Table~\ref{tab:quant_tikz} shows that the gains from intermediate-render
supervision also extend to Image-to-TikZ generation. Our method improves
over the DeTikZify-v2 SFT base model and over outcome-only RL across all metrics, while once again producing shorter TikZ programs. 
Figure~\ref{fig:image_to_tikz_qual} shows representative qualitative
comparisons and reflects the same trend. Our method more faithfully
reconstructs the visual structure and content of the target figures,
while the supervised model and outcome-based GRPO more frequently omit or distort visual elements.

Across both SVG and TikZ, our method outperforms SFT and outcome-only RL, demonstrating its effectiveness across distinct image-to-code representations.

\subsection{Analysis}
\label{sec:analysis}

\begin{figure*}[t!]
  \centering
  \includegraphics[width=\linewidth]{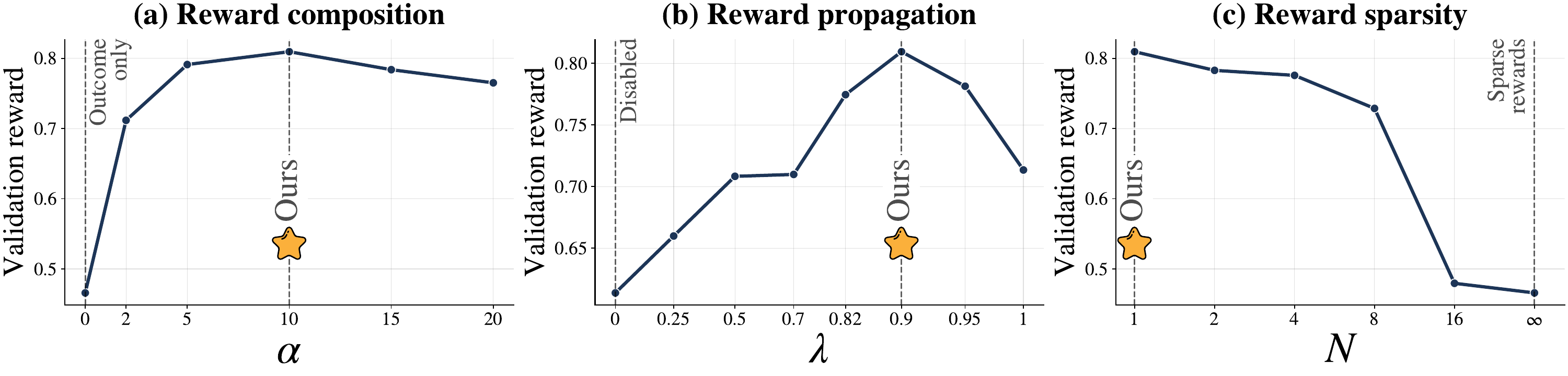}
  \caption{ \textbf{Ablations of process supervision on SVG-Stack validation.} (a) Increasing the process contribution improves performance up to $\alpha=10$. (b) Reward propagation performs best at $\lambda=0.9$. (c) More frequent intermediate renders consistently improve performance. Stars mark our selected settings, which also perform best on our test set MMSVGBench (Fig.~\ref{fig:analysis_extended}).}
  \label{fig:analysis}
\end{figure*}

\begin{figure}[t!]
  \centering
  \includegraphics[width=1.0\linewidth, keepaspectratio]{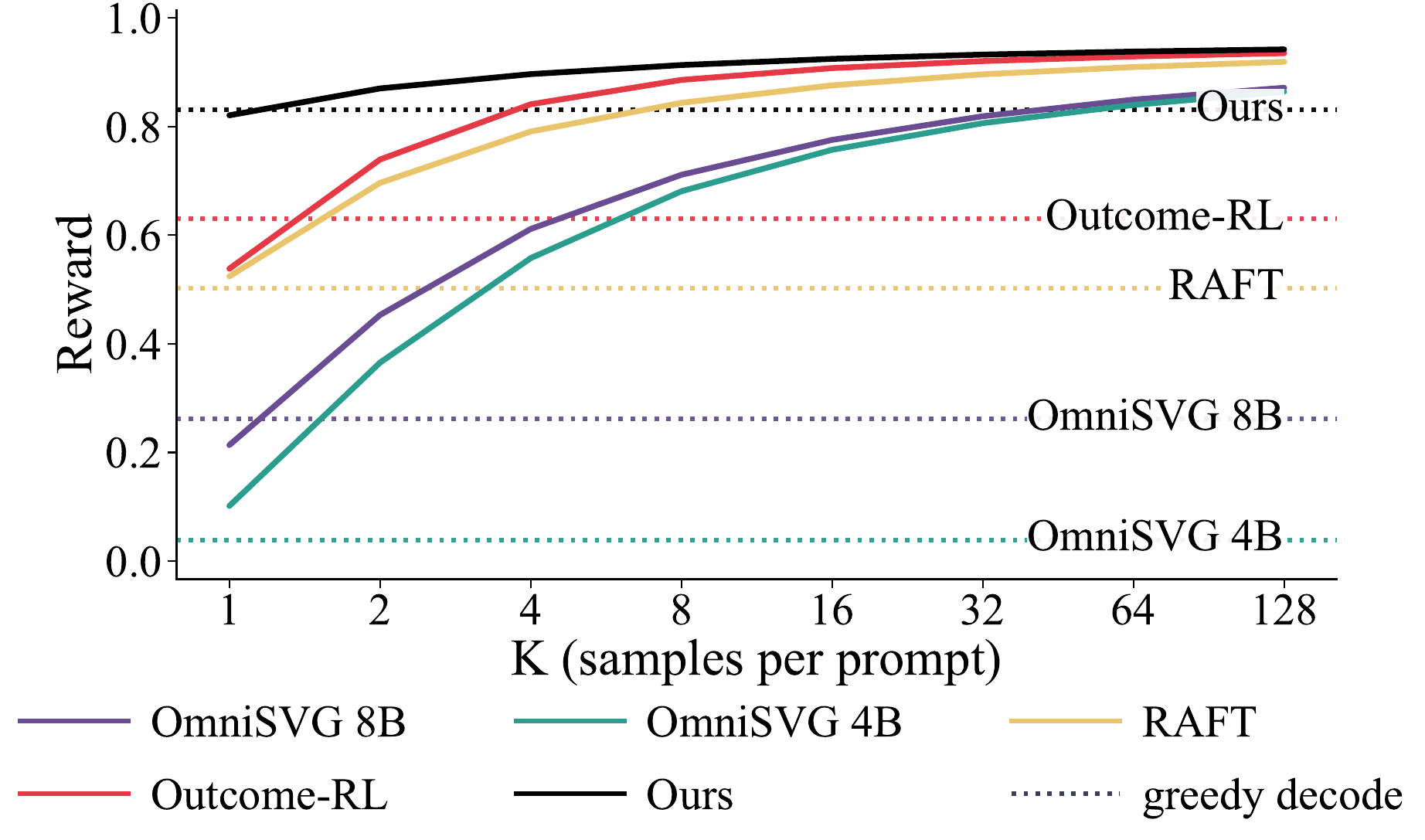}
    \caption{
    \textbf{Best-of-$K$ sampling.}
    For each image, we sample $K$ candidate SVGs and report the mean reward of the best ones. Horizontal lines show greedy decoding. Our method performs best, with the largest gains at small $K$.
    }
    \label{fig:best_of_k}
    
\end{figure}

We study the contributions of process and outcome rewards, the impact of reward propagation and intermediate rendering frequency, training efficiency, and test-time sampling performance, reporting Image-to-SVG results below and Image-to-TikZ results in App.~\ref{sec:ext_analysis}.

\vspace{-10pt}
\paragraph{Process and Outcome Rewards.} 
We first ablate the two components of our objective
(Eq.~\ref{eq:combined_advantage}). Process-only supervision outperforms outcome-only,  and combining both performs best across all metrics (Tab.~\ref{tab:quant}). Fig.~\ref{fig:reward_ablation} shows the same qualitative
trend. We report corresponding TikZ ablation in Tab.~\ref{tab:ablation_tikz}.
We further vary the weight of the process reward $\alpha$ (Eq.~\ref{eq:combined_advantage}) at Fig.~\ref{fig:analysis}a. Increasing its contribution peaks the validation reward at $\alpha=10$. Larger values provide no further gain.

\vspace{-10pt}
\paragraph{Reward Propagation.}
We vary $\lambda$, which controls how far each render-progress reward propagates to preceding tokens (Eq.~\ref{eq:progress_credit}), and report results in Fig.~\ref{fig:analysis}b. Without discounting ($\lambda=1$), each token is credited with the total visual improvement from its position onward. This is equivalent to the outcome reward with a per-prefix baseline given by the intermediate render. We note it outperforms outcome-only training ($\alpha=0$).
Discounting ($\lambda<1$) further biases credit toward nearby render changes, with performance peaking at $\lambda=0.9$. We discuss this bias--variance trade-off in the context of GAE in App.~\ref{sec:pbrs_gae}.

\vspace{-10pt}
\paragraph{Render Frequency.} Our default formulation renders after every completed drawing command (Sec.~\ref{sec:progress}). We vary this frequency by rendering once every $N$ commands. 
Figure~\ref{fig:analysis}c shows that performance decreases as renders become less frequent, with best result at $N=1$. This supports measuring progress as frequently as the representation allows.

In App.~\ref{sec:ext_analysis} we extend this analysis to Image-to-TikZ and a broader set of metrics. The same conclusions hold with similar optimal settings, suggesting reasonable defaults for new tasks.

\begin{figure}[t]
  \centering
  \includegraphics[width=\linewidth, keepaspectratio]{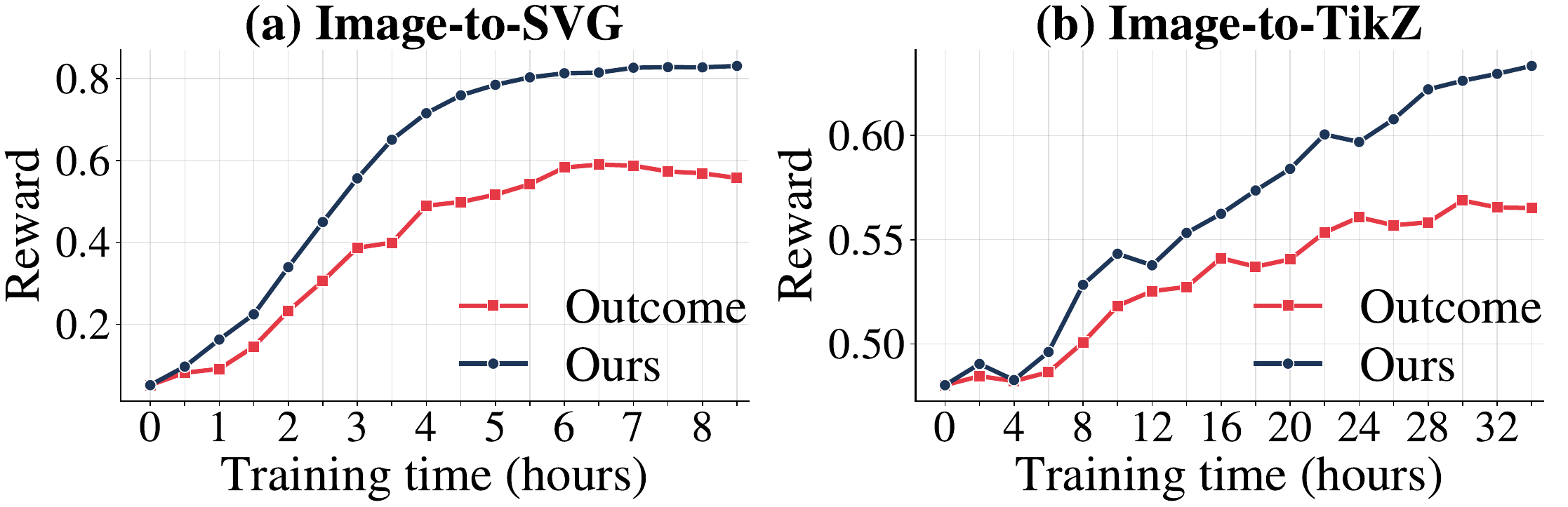}
    \caption{
    \textbf{Training dynamics.} Despite the extra cost of process rewards, our method outperforms outcome-only training in test reward even at equal training time.
    }
    \label{fig:converge}
\end{figure}

\vspace{-10pt}
\paragraph{Training efficiency.}
Process reward computation takes roughly half of each training step's runtime, as shown in the runtime analysis in App.~\ref{sec:ext_analysis}.
However, even at equal wall-clock time, our method consistently outperforms outcome-only training on both tasks (Fig.~\ref{fig:converge}) and on almost all test metrics (Tab.~\ref{tab:metrics_runtime}), justifying the extra per-step cost.

\paragraph{Test-time Scaling.}

We evaluate how sampling of additional $K$ candidates per image improves generation quality.
Figure~\ref{fig:best_of_k} reports the average best-of-$K$ reward over the Icons and Illustrations subsets of MMSVGBench, showing that our model consistently achieves the highest reward, with the largest margin at small $K$. 
The narrowing gap at larger $K$ suggests that training primarily increases the probability of high-quality generations of the base model, consistent with prior observations ~\citep{yue2025limit}.
In App.~\ref{sec:test_time_extended}, we provide separate results for MMSVGBench subsets, extend evaluation to SVG-Stack, and consider additional sampling temperatures, observing the same trend.
Notably, OmniSVG-8B is much stronger than OmniSVG-4B at $K=1$, yet the gap nearly vanishes at large $K$. We test the impact of the base model on our method, observing both models reach similar quality (App.~\ref{sec:svg_model_scale}).


\section{Conclusions}
We introduced \method, a method for reinforcement learning from intermediate renders for
image-to-code generation. By using changes between consecutive renders,
our method provides localized reward during generation while retaining
the global outcome signal. Across Image-to-SVG and Image-to-TikZ, it
consistently improves over supervised fine-tuning and outcome-only RL,
yielding state-of-the-art results among open-source models. Our analysis further shows that process and outcome feedback are complementary, with their combination performing best, and that denser intermediate renders yield stronger supervision. 

A limitation is that intermediate program prefixes must be convertible
into meaningful executable states that can be evaluated against the
target. 
We show that it holds naturally for Image-to-SVG and Image-to-TikZ tasks. Future work  could extend the idea to
other incrementally rendered representations, such as Lottie animations,
HTML interfaces, and 3D scene programs. 
As base models continue to advance rapidly through large-scale pretraining,
our method could serve as a natural post-training step for image-to-code
tasks. More broadly, our
results highlight intermediate execution as a natural source of
process-level supervision for improving credit assignment beyond terminal
rewards.

\section*{LLM Usage Statement}
In this work, we used LLMs (ChatGPT Astra/Sol, and Claude Opus 5.5) to implement methods, creating and editing software code, drafting parts of the paper, editing for readability, identifying relevant literature, and formatting references. We reviewed all LLM-assisted work: code was tested and verified for correctness by the authors, and all LLM-drafted or LLM-edited text was checked against the underlying results and revised by the authors. We take responsibility for the final content of this work, including text, claims, and artifacts produced with the aid of LLM.





\section*{Acknowledgments}
This research was supported by the Sagol Weizmann-MIT Bridge Program and made possible through a GPU compute resource grant funded by the Association of University Heads, the Council for Higher Education and the AI Research Compute Center.

{
    \small
    \bibliographystyle{ieeenat_fullname}
    \bibliography{main}

@String(CVPR= {IEEE Conf. Comput. Vis. Pattern Recog.})

@String(ICCV= {Int. Conf. Comput. Vis.})

@String(ICLR = {Int. Conf. Learn. Represent.})

@String(CVPR  = {CVPR})

@String(ICCV  = {ICCV})

@String(ICLR  = {ICLR})

@inproceedings{yang2025omnisvg,
  title={OmniSVG: A Unified Scalable Vector Graphics Generation Model},
  author={Yang, Yiying and Cheng, Wei and Chen, Sijin and Zeng, Xianfang and Yin, Fukun and Zhang, Jiaxu and Wang, Liao and Yu, Gang and Ma, Xingjun and Jiang, Yu-Gang},
  booktitle={Advances in Neural Information Processing Systems},
  volume={38},
  year={2025},
  doi={10.52202/085713-3791}
}

@inproceedings{belouadi2024detikzify,
    title={{DeTikZify}: Synthesizing Graphics Programs for Scientific Figures and Sketches with {TikZ}},
    author={Jonas Belouadi and Simone Paolo Ponzetto and Steffen Eger},
    booktitle={The Thirty-eighth Annual Conference on Neural Information Processing Systems},
    year={2024},
    url={https://openreview.net/forum?id=bcVLFQCOjc}
}

@article{zhao2025vincicoder,
  title={Vincicoder: Unifying multimodal code generation via coarse-to-fine visual reinforcement learning},
  author={Zhao, Xuanle and Jiang, Deyang and Zeng, Zhixiong and Chen, Lei and Qiu, Haibo and Huang, Jing and Zhong, Yufeng and Zheng, Liming and Cao, Yilin and Ma, Lin},
  journal={arXiv preprint arXiv:2511.00391},
  year={2025}
}

@article{yang2026omnilottie,
  title={OmniLottie: Generating Vector Animations via Parameterized Lottie Tokens},
  author={Yiying Yang and Wei Cheng and Sijin Chen and Honghao Fu and Xianfang Zeng and Yujun Cai and Gang Yu and Xinjun Ma},
  journal={arXiv preprint arxiv:2603.02138},
  year={2026}
}

@InProceedings{Chen_2026_CVPR_lottiegpt,
  author    = {Chen, Junhao and Gao, Kejun and Cui, Yuehan and Sun, Mingze and Chen, Mingjin and Wang, Shaohui and Long, Xiaoxiao and Ma, Fei and Tian, Qi and Zhao, Hao and Huang, Ruqi},
  title     = {LottieGPT: Tokenizing Vector Animation for Autoregressive Generation},
  booktitle = {Proceedings of the IEEE/CVF Conference on Computer Vision and Pattern Recognition (CVPR)},
  month     = {June},
  year      = {2026},
  pages     = {31639-31651}
}

@inproceedings{chen2025cadcrafter,
  title={Cadcrafter: Generating computer-aided design models from unconstrained images},
  author={Chen, Cheng and Wei, Jiacheng and Chen, Tianrun and Zhang, Chi and Yang, Xiaofeng and Zhang, Shangzhan and Yang, Bingchen and Foo, Chuan-Sheng and Lin, Guosheng and Huang, Qixing and others},
  booktitle={2025 IEEE/CVF Conference on Computer Vision and Pattern Recognition (CVPR)},
  pages={11073--11082},
  year={2025},
  organization={IEEE}
}

@inproceedings{rodriguez2026rendering_rlrf,
  title={Rendering-aware reinforcement learning for vector graphics generation},
  author={Rodriguez, Juan A. and Zhang, Haotian and Puri, Abhay and Feizi, Aarash and Pramanik, Rishav and Wichmann, Pascal and Mondal, Arnab and Samsami, Mohammad Reza and Awal, Rabiul and Taslakian, Perouz and Gella, Spandana and Rajeswar, Sai and Vazquez, David and Pal, Christopher and Pedersoli, Marco},
  booktitle={Advances in Neural Information Processing Systems},
  volume={38},
  pages={67317--67355},
  year={2025}
}

@misc{deepseek-math,
  author = {Shao, Zhihong and Wang, Peiyi and Zhu, Qihao and Xu, Runxin and Song, Junxiao and Bi, Xiao and Zhang, Haowei and Zhang, Mingchuan and Li, Y. K. and Wu, Y. and Guo, Daya},
  title = {DeepSeekMath: Pushing the Limits of Mathematical Reasoning in Open Language Models},
  year = {2024},
  eprint = {2402.03300},
  archivePrefix = {arXiv},
  url = {https://arxiv.org/abs/2402.03300},
}

@article{schulman2015high,
  title={High-dimensional continuous control using generalized advantage estimation},
  author={Schulman, John and Moritz, Philipp and Levine, Sergey and Jordan, Michael and Abbeel, Pieter},
  journal={arXiv preprint arXiv:1506.02438},
  year={2015}
}

@inproceedings{rodriguez2023starvector,
    title={{StarVector: Generating Scalable Vector Graphics Code from Images and Text}},
    author={Juan A. Rodriguez and Abhay Puri and Shubham Agarwal and Issam H. Laradji and Pau Rodriguez and Sai Rajeswar and David Vazquez and Christopher Pal and Marco Pedersoli},
    booktitle={Proceedings of the IEEE/CVF Conference on Computer Vision and Pattern Recognition (CVPR)},
    month={June},
    year={2025},
    pages={16175--16186},
}

@article{dong2023raft,
  title={{RAFT}: Reward rAnked FineTuning for Generative Foundation Model Alignment},
  author={Hanze Dong and Wei Xiong and Deepanshu Goyal and Yihan Zhang and Winnie Chow and Rui Pan and Shizhe Diao and Jipeng Zhang and KaShun SHUM and Tong Zhang},
  journal={Transactions on Machine Learning Research},
  issn={2835-8856},
  year={2023},
  url={https://openreview.net/forum?id=m7p5O7zblY},
}

@inproceedings{wang2025internsvg,
  title={InternSVG: Towards Unified SVG Tasks with Multimodal Large Language Models},
  author={Wang, Haomin and Yin, Jinhui and Wei, Qi and Zeng, Wenguang and Gu, Lixin and Ye, Shenglong and Gao, Zhangwei and Wang, Yaohui and Zhang, Yanting and Li, Yuanqi and Guo, Yanwen and Wang, Wenhai and Chen, Kai and Qiao, Yu and Zhang, Hongjie},
  booktitle={International Conference on Learning Representations},
  year={2026},
  eprint={2510.11341},
  archivePrefix={arXiv}
}

@article{DiffVG,
    title = {Differentiable Vector Graphics Rasterization for Editing and Learning},
    author = {Li, Tzu-Mao and Luk\'{a}\v{c}, Michal and Gharbi, Micha\"{e}l and Ragan-Kelley, Jonathan},
    journal = {ACM Trans. Graph. (Proc. SIGGRAPH Asia)},
    volume = {39},
    number = {6},
    pages = {193:1--193:15},
    year = {2020}
}

@inproceedings{xu2022live,
    title={Towards Layer-wise Image Vectorization},
    author={Ma, Xu and Zhou, Yuqian and Xu, Xingqian and Sun, Bin and Filev, Valerii and  Orlov, Nikita and Fu, Yun and Shi, Humphrey},
    booktitle={Proceedings of the IEEE conference on computer vision and pattern recognition},
    year={2022}
}

@article{Qwen3-VL,
      title={Qwen3-VL Technical Report}, 
      author={Shuai Bai and Yuxuan Cai and Ruizhe Chen and Keqin Chen and Xionghui Chen and Zesen Cheng and Lianghao Deng and Wei Ding and Chang Gao and Chunjiang Ge and Wenbin Ge and Zhifang Guo and Qidong Huang and Jie Huang and Fei Huang and Binyuan Hui and Shutong Jiang and Zhaohai Li and Mingsheng Li and Mei Li and Kaixin Li and Zicheng Lin and Junyang Lin and Xuejing Liu and Jiawei Liu and Chenglong Liu and Yang Liu and Dayiheng Liu and Shixuan Liu and Dunjie Lu and Ruilin Luo and Chenxu Lv and Rui Men and Lingchen Meng and Xuancheng Ren and Xingzhang Ren and Sibo Song and Yuchong Sun and Jun Tang and Jianhong Tu and Jianqiang Wan and Peng Wang and Pengfei Wang and Qiuyue Wang and Yuxuan Wang and Tianbao Xie and Yiheng Xu and Haiyang Xu and Jin Xu and Zhibo Yang and Mingkun Yang and Jianxin Yang and An Yang and Bowen Yu and Fei Zhang and Hang Zhang and Xi Zhang and Bo Zheng and Humen Zhong and Jingren Zhou and Fan Zhou and Jing Zhou and Yuanzhi Zhu and Ke Zhu},
	  journal={arXiv preprint arXiv:2511.21631},
      year={2025}
}

@misc{gemini3flash,
      title={Gemini 3 Flash},
      author={{Google DeepMind}},
      year={2025},
      month={12},
      howpublished={\url{https://deepmind.google/models/gemini/flash/}},
      note={Model card. Accessed 2026-09-22}
}

@misc{claudesonnet5,
      title={Claude Sonnet 5 System Card},
      author={{Anthropic}},
      year={2026},
      month={6},
      howpublished={\url{https://anthropic.com/claude-sonnet-5-system-card}},
      note={System card. Accessed 2026-09-22}
}

@misc{gpt52,
      title={GPT-5.2 System Card},
      author={{OpenAI}},
      year={2025},
      month={12},
      howpublished={\url{https://openai.com/index/gpt-5-system-card-update-gpt-5-2}},
      note={System card. Accessed 2026-09-22}
}

@misc{tikzero2025,
  title={{TikZero}: Zero-Shot Text-Guided Graphics Program Synthesis},
  author={Belouadi, Jonas and Ilg, Eddy and Keuper, Margret and Tanaka, Hideki and Utiyama, Masao and Dabre, Raj and Eger, Steffen and Ponzetto, Simone Paolo},
  year={2025},
  eprint={2503.11509},
  archivePrefix={arXiv},
  url={https://arxiv.org/abs/2503.11509}
}

@inproceedings{yun2024web2code,
  title     = {{Web2Code}: A Large-scale Webpage-to-Code Dataset and Evaluation Framework for Multimodal {LLMs}},
  author    = {Yun, Sukmin and Lin, Haokun and Thushara, Rusiru and Bhat, Mohammad Qazim and Wang, Yongxin and Jiang, Zutao and Deng, Mingkai and Wang, Jinhong and Tao, Tianhua and Li, Junbo and Li, Haonan and Nakov, Preslav and Baldwin, Timothy and Liu, Zhengzhong and Xing, Eric P. and Liang, Xiaodan and Shen, Zhiqiang},
  booktitle = {Advances in Neural Information Processing Systems},
  volume    = {37},
  pages     = {112134--112157},
  year      = {2024},
  doi       = {10.52202/079017-3560}
}

@article{chen2025img2cad,
  title   = {{Img2CAD}: Conditioned {3D} {CAD} Model Generation from Single Image with Structured Visual Geometry},
  author  = {Chen, Tianrun and Yu, Chunan and Hu, Yuanqi and Li, Jing and Xu, Tao and Cao, Runlong and Zhu, Lanyun and Zang, Ying and Zhang, Yong and Li, Zejian and Sun, Lingyun},
  journal = {IEEE Transactions on Industrial Informatics},
  volume  = {21},
  number  = {11},
  pages   = {8539--8549},
  year    = {2025}
}

@article{zhao2026beyondnl2code,
  title         = {Beyond {NL2Code}: A Structured Survey of Multimodal Code Intelligence},
  author        = {Zhao, Xuanle and Sun, Qiushi and Xiao, Jingyu and Liu, Xuexin and Yang, Haoyue and Chen, Qiaosheng and Luo, Xianzhen and Huang, Jing and Zhong, Yufeng and Chen, Lei and Fu, Shuai and Wei, Zhenlin and Bi, Jinhe and Jiang, Lei and Qiu, Haibo and Yang, Siqi and Shi, Peng and Hu, Jian and Zeng, Zhixiong},
  journal       = {arXiv preprint arXiv:2606.15932},
  year          = {2026},
  eprint        = {2606.15932},
  archivePrefix = {arXiv},
  url           = {https://arxiv.org/abs/2606.15932}
}

@misc{tan2025chartmaster,
  title         = {{ChartMaster}: Advancing Chart-to-Code Generation with Real-World Charts and Chart Similarity Reinforcement Learning},
  author        = {Tan, Wentao and Cao, Qiong and Xue, Chao and Zhan, Yibing and Ding, Changxing and He, Xiaodong},
  year          = {2025},
  eprint        = {2508.17608},
  archivePrefix = {arXiv},
  url           = {https://arxiv.org/abs/2508.17608}
}

@inproceedings{ling2025table2latexrl,
  title     = {{Table2LaTeX-RL}: High-Fidelity {LaTeX} Code Generation from Table Images via Reinforced Multimodal Language Models},
  author    = {Ling, Jun and Qi, Yao and Huang, Tao and Zhou, Shibo and Huang, Yanqin and Yang, Jiang and Song, Ziqi and Zhou, Ying and Yang, Yang and Shen, Heng Tao and Wang, Peng},
  booktitle = {Advances in Neural Information Processing Systems},
  volume    = {38},
  year      = {2025}
}

@inproceedings{saito2025sketch2diagram,
  title     = {{Sketch2Diagram}: Generating Vector Diagrams from Hand-Drawn Sketches},
  author    = {Saito, Itsumi and Yoshida, Haruto and Sakaguchi, Keisuke},
  booktitle = {International Conference on Learning Representations},
  year      = {2025}
}

@misc{wang2026ctrls,
  title         = {Reliable Reasoning in {SVG-LLMs} via Multi-Task Multi-Reward Reinforcement Learning},
  author        = {Wang, Haomin and Wei, Qi and Ma, Qianli and Ding, Shengyuan and Yin, Jinhui and Chen, Kai and Zhang, Hongjie},
  year          = {2026},
  eprint        = {2603.16189},
  archivePrefix = {arXiv},
  url           = {https://arxiv.org/abs/2603.16189}
}

@inproceedings{zeng2026davinci,
  title     = {{DaVinci}: Reinforcing Visual-Structural Syntax in {MLLMs} for Generalized Scientific Diagram Parsing},
  author    = {Zeng, Xingchen and Su, Zhewei and Zhang, Hengming and Jiang, Juyong and Xia, Jiazhi and Zeng, Wei},
  booktitle = {International Conference on Learning Representations},
  year      = {2026}
}

@misc{liang2026renderintheloop,
  title         = {Render-in-the-Loop: Vector Graphics Generation via Visual Self-Feedback},
  author        = {Liang, Guotao and Wang, Zhangcheng and Hu, Juncheng and Zhou, Haitao and Xue, Ziteng and Zhang, Jing and Xu, Dong and Yu, Qian},
  year          = {2026},
  eprint        = {2604.20730},
  archivePrefix = {arXiv},
  url           = {https://arxiv.org/abs/2604.20730}
}

@misc{deng2026visrefiner,
  title         = {{VisRefiner}: Learning from Visual Differences for Screenshot-to-Code Generation},
  author        = {Deng, Jie and Yao, Kaichun and Zhang, Libo},
  year          = {2026},
  eprint        = {2602.05998},
  archivePrefix = {arXiv},
  url           = {https://arxiv.org/abs/2602.05998}
}

@inproceedings{yang2026ui2coden,
  title         = {{UI2Code$^{\mathrm{N}}$}: {UI}-to-Code Generation as Interactive Visual Optimization},
  author        = {Yang, Zhen and Hong, Wenyi and Xu, Mingde and Fan, Xinyue and Wang, Weihan and Cheng, Jiale and Gu, Xiaotao and Tang, Jie},
  booktitle     = {Proceedings of the 43rd International Conference on Machine Learning},
  series        = {Proceedings of Machine Learning Research},
  volume        = {306},
  year          = {2026},
  eprint        = {2511.08195},
  archivePrefix = {arXiv},
  url           = {https://arxiv.org/abs/2511.08195}
}

@misc{liu2026visualerm,
  title         = {Visual-{ERM}: Reward Modeling for Visual Equivalence},
  author        = {Liu, Ziyu and Ding, Shengyuan and Fang, Xinyu and Dai, Xuanlang and Yang, Penghui and Liang, Jianze and Wang, Jiaqi and Chen, Kai and Lin, Dahua and Zang, Yuhang},
  year          = {2026},
  eprint        = {2603.13224},
  archivePrefix = {arXiv},
  url           = {https://arxiv.org/abs/2603.13224}
}

@misc{prolific2024,
  author       = {Prolific},
  title        = {Prolific},
  year         = {2024},
  howpublished = {\url{https://www.prolific.com/}}
}

@inproceedings{yang2021neurips-tuning,
  title     = {Tuning Large Neural Networks via Zero-Shot Hyperparameter Transfer},
  author    = {Yang, Ge and Hu, Edward and Babuschkin, Igor and Sidor, Szymon and Liu, Xiaodong and Farhi, David and Ryder, Nick and Pachocki, Jakub and Chen, Weizhu and Gao, Jianfeng},
  booktitle = {Advances in Neural Information Processing Systems},
  year      = {2021}
}

@misc{cobbe2021training,
      title={Training Verifiers to Solve Math Word Problems},
      author={Karl Cobbe and Vineet Kosaraju and Mohammad Bavarian and Jacob Hilton and Reiichiro Nakano and Christopher Hesse and John Schulman},
      year={2021},
      eprint={2110.14168},
      archivePrefix={arXiv},
      primaryClass={cs.LG}
}

@article{uesato2022solving,
    title        = {Solving Math Word Problems With Process- and Outcome-Based Feedback},
    author       = {Uesato, Jonathan and Kushman, Nate and Kumar, Ramana and Song, Francis and Siegel, Noah and Wang, Lisa and Creswell, Antonia and Irving, Geoffrey and Higgins, Irina},
    year         = 2022,
    journal      = {arXiv preprint arXiv:2211.14275}
}

@inproceedings{lightman2024verify,
  title={Let's verify step by step},
  author={Lightman, Hunter and Kosaraju, Vineet and Burda, Yuri and Edwards, Harrison and Baker, Bowen and Lee, Teddy and Leike, Jan and Schulman, John and Sutskever, Ilya and Cobbe, Karl},
  booktitle={International Conference on Learning Representations},
  volume={2024},
  pages={39578--39601},
  year={2024}
}

@inproceedings{wang2024mathshepherd,
  title={Math-shepherd: Verify and reinforce llms step-by-step without human annotations},
  author={Wang, Peiyi and Li, Lei and Shao, Zhihong and Xu, Runxin and Dai, Damai and Li, Yifei and Chen, Deli and Wu, Yu and Sui, Zhifang},
  booktitle={Proceedings of the 62nd Annual Meeting of the Association for Computational Linguistics (Volume 1: Long Papers)},
  pages={9426--9439},
  year={2024}
}

@inproceedings{setlur2025rewarding,
  title={Rewarding progress: Scaling automated process verifiers for llm reasoning},
  author={Setlur, Amrith and Nagpal, Chirag and Fisch, Adam and Geng, Xinyang and Eisenstein, Jacob and Agarwal, Rishabh and Agarwal, Alekh and Berant, Jonathan and Kumar, Aviral},
  booktitle={International Conference on Learning Representations},
  volume={2025},
  pages={60808--60838},
  year={2025}
}

@article{kim2026process,
  title={Process-verified reinforcement learning for theorem proving via Lean},
  author={Kim, Minsu and Yun, Se-Young},
  journal={arXiv preprint arXiv:2606.20068},
  year={2026}
}

@inproceedings{dou2024stepcoder,
  title={Stepcoder: improving code generation with reinforcement learning from compiler feedback},
  author={Dou, Shihan and Liu, Yan and Jia, Haoxiang and Zhou, Enyu and Xiong, Limao and Shan, Junjie and Huang, Caishuang and Wang, Xiao and Fan, Xiaoran and Xi, Zhiheng and others},
  booktitle={Proceedings of the 62nd Annual Meeting of the Association for Computational Linguistics (Volume 1: Long Papers)},
  pages={4571--4585},
  year={2024}
}

@inproceedings{ye2025process,
  title={Process-supervised reinforcement learning for code generation},
  author={Ye, Yufan and Zhang, Ting and Jiang, Wenbin and Huang, Hua},
  booktitle={Proceedings of the 2025 Conference on Empirical Methods in Natural Language Processing},
  pages={14224--14237},
  year={2025}
}

@inproceedings{yu2024rlhf,
  title={Rlhf-v: Towards trustworthy mllms via behavior alignment from fine-grained correctional human feedback},
  author={Yu, Tianyu and Yao, Yuan and Zhang, Haoye and He, Taiwen and Han, Yifeng and Cui, Ganqu and Hu, Jinyi and Liu, Zhiyuan and Zheng, Hai-Tao and Sun, Maosong and others},
  booktitle={Proceedings of the IEEE/CVF Conference on Computer Vision and Pattern Recognition},
  pages={13807--13816},
  year={2024}
}

@article{yue2025limit,
  title={Does Reinforcement Learning Really Incentivize Reasoning Capacity in LLMs Beyond the Base Model?},
  author={Yue, Yang and Chen, Zhiqi and Lu, Rui and Zhao, Andrew and Wang, Zhaokai and Yue, Yang and Song, Shiji and Huang, Gao},
  journal={arXiv preprint arXiv:2504.13837},
  year={2025}
}

@article{oquab2024dinov2,
  title   = {{DINOv2}: Learning Robust Visual Features without Supervision},
  author  = {Oquab, Maxime and Darcet, Timoth{\'e}e and Moutakanni, Th{\'e}o and Vo, Huy V. and Szafraniec, Marc and Khalidov, Vasil and Fernandez, Pierre and Haziza, Daniel and Massa, Francisco and El-Nouby, Alaaeldin and Assran, Mahmoud and Ballas, Nicolas and Galuba, Wojciech and Howes, Russell and Huang, Po-Yao and Li, Shang-Wen and Misra, Ishan and Rabbat, Michael and Sharma, Vasu and Synnaeve, Gabriel and Xu, Hu and Jegou, Herv{\'e} and Mairal, Julien and Labatut, Patrick and Joulin, Armand and Bojanowski, Piotr},
  journal = {Transactions on Machine Learning Research},
  year    = {2024},
  note    = {arXiv:2304.07193}
}

@article{wang2004ssim,
  title   = {Image Quality Assessment: From Error Visibility to Structural Similarity},
  author  = {Wang, Zhou and Bovik, Alan C. and Sheikh, Hamid R. and Simoncelli, Eero P.},
  journal = {IEEE Transactions on Image Processing},
  volume  = {13},
  number  = {4},
  pages   = {600--612},
  year    = {2004},
  doi     = {10.1109/TIP.2003.819861}
}

@inproceedings{zhang2018lpips,
  title     = {The Unreasonable Effectiveness of Deep Features as a Perceptual Metric},
  author    = {Zhang, Richard and Isola, Phillip and Efros, Alexei A. and Shechtman, Eli and Wang, Oliver},
  booktitle = {Proceedings of the IEEE Conference on Computer Vision and Pattern Recognition (CVPR)},
  pages     = {586--595},
  year      = {2018}
}

@inproceedings{radford2021clip,
  title     = {Learning Transferable Visual Models From Natural Language Supervision},
  author    = {Radford, Alec and Kim, Jong Wook and Hallacy, Chris and Ramesh, Aditya and Goh, Gabriel and Agarwal, Sandhini and Sastry, Girish and Askell, Amanda and Mishkin, Pamela and Clark, Jack and Krueger, Gretchen and Sutskever, Ilya},
  booktitle = {Proceedings of the 38th International Conference on Machine Learning (ICML)},
  series    = {Proceedings of Machine Learning Research},
  volume    = {139},
  pages     = {8748--8763},
  year      = {2021}
}

@inproceedings{hu2022lora,
    title={Lo{RA}: Low-Rank Adaptation of Large Language Models},
    author={Edward J Hu and Yelong Shen and Phillip Wallis and Zeyuan Allen-Zhu and Yuanzhi Li and Shean Wang and Lu Wang and Weizhu Chen},
    booktitle={International Conference on Learning Representations},
    year={2022},
    url={https://openreview.net/forum?id=nZeVKeeFYf9}
  }

@article{wang2025reinforcement,
    title={Reinforcement Learning for Reasoning in Large Language Models with One Training Example},
    author={Wang, Yiping and Yang, Qing and Zeng, Zhiyuan and Ren, Liliang and Liu, Liyuan and Peng, Baolin and Cheng, Hao and He, Xuehai and Wang, Kuan and Gao, Jianfeng and Chen, Weizhu and Wang, Shuohang and
  Du, Simon Shaolei and Shen, Yelong},
    journal={arXiv preprint arXiv:2504.20571},
    year={2025}
  }

@inproceedings{zhai2023sigmoid,
  title     = {Sigmoid Loss for Language Image Pre-Training},
  author    = {Zhai, Xiaohua and Mustafa, Basil and Kolesnikov, Alexander and Beyer, Lucas},
  booktitle = {Proceedings of the IEEE/CVF International Conference on Computer Vision (ICCV)},
  pages     = {11975--11986},
  year      = {2023}
}

@inproceedings{fu2023dreamsim,
  title     = {DreamSim: Learning New Dimensions of Human Visual Similarity using Synthetic Data},
  author    = {Fu, Stephanie and Tamir, Netanel Y. and Sundaram, Shobhita and Chai, Lucy and Zhang, Richard and Dekel, Tali and Isola, Phillip},
  booktitle = {Advances in Neural Information Processing Systems (NeurIPS)},
  year      = {2023}
}

@inproceedings{rombach2022high,
  title={High-resolution image synthesis with latent diffusion models},
  author={Rombach, Robin and Blattmann, Andreas and Lorenz, Dominik and Esser, Patrick and Ommer, Bj{\"o}rn},
  booktitle={2022 IEEE/CVF conference on computer vision and pattern recognition (CVPR)},
  pages={10674--10685},
  year={2022},
  organization={ieee}
}

@inproceedings{esser2024scaling,
  title={Scaling rectified flow transformers for high-resolution image synthesis},
  author={Esser, Patrick and Kulal, Sumith and Blattmann, Andreas and Entezari, Rahim and M{\"u}ller, Jonas and Saini, Harry and Levi, Yam and Lorenz, Dominik and Sauer, Axel and Boesel, Frederic and others},
  booktitle={Forty-first international conference on machine learning},
  year={2024}
}

@article{wu2023human,
    title={Human Preference Score v2: A Solid Benchmark for Evaluating Human Preferences of Text-to-Image Synthesis},
    author={Wu, Xiaoshi and Hao, Yiming and Sun, Keqiang and Chen, Yixiong and Zhu, Feng and Zhao, Rui and Li, Hongsheng},
    journal={arXiv preprint arXiv:2306.09341},
    year={2023}
  }

@inproceedings{ng1999policy,
  title={Policy invariance under reward transformations: Theory and application to reward shaping},
  author={Ng, Andrew Y. and Harada, Daishi and Russell, Stuart},
  booktitle={Proceedings of the Sixteenth International Conference on Machine Learning (ICML)},
  pages={278--287},
  year={1999}
}

@article{williams1992simple,
  title   = {Simple Statistical Gradient-Following Algorithms for Connectionist Reinforcement Learning},
  author  = {Williams, Ronald J.},
  journal = {Machine Learning},
  volume  = {8},
  number  = {3--4},
  pages   = {229--256},
  year    = {1992}
}

@inproceedings{loshchilov2019decoupled,
  title     = {Decoupled Weight Decay Regularization},
  author    = {Loshchilov, Ilya and Hutter, Frank},
  booktitle = {International Conference on Learning Representations (ICLR)},
  year      = {2019}
}

@inproceedings{huang2019learning,
  title={Learning to paint with model-based deep reinforcement learning},
  author={Huang, Zhewei and Heng, Wen and Zhou, Shuchang},
  booktitle={Proceedings of the IEEE/CVF international conference on computer vision},
  pages={8709--8718},
  year={2019}
}

@article{bai2025qwen25vl,
    title   = {Qwen2.5-VL Technical Report},
    author  = {Bai, Shuai and Chen, Keqin and Liu, Xuejing and Wang, Jialin and Ge, Wenbin and Song, Sibo and Dang, Kai and Wang, Peng and Wang, Shijie and Tang, Jun and Zhong, Humen and Zhu, Yuanzhi and
  Yang, Mingkun and Li, Zhaohai and Wan, Jianqiang and Wang, Pengfei and Ding, Wei and Fu, Zheren and Xu, Yiheng and Ye, Jiabo and Zhang, Xi and Xie, Tianbao and Cheng, Zesen and Zhang, Hang and Yang,
  Zhibo and Xu, Haiyang and Lin, Junyang},
    journal = {arXiv preprint arXiv:2502.13923},
    year    = {2025}
  }

@misc{detikzify_v2_5_8b,
  author       = {Belouadi, Jonas},
  title        = {{DeTikZify-v2.5 (8b)}},
  year         = {2025},
  howpublished = {Hugging Face model card},
  url          = {https://huggingface.co/nllg/detikzify-v2.5-8b},
  note         = {Accessed 2026-11-24}
}
}
\clearpage

\appendix

\counterwithin{figure}{section}
\counterwithin{table}{section}
\counterwithin{equation}{section}

{\Large\centering\textbf{Appendix}\par}

\begingroup
  \renewcommand{\sectionmark}[1]{}
  \renewcommand{\subsectionmark}[1]{}
  \noindent\textbf{Appendix Contents}\\[0.5em]
  \hyperref[sec:pbrs_gae]{A}\quad \textbf{Connection to PBRS and GAE}\dotfill\pageref{sec:pbrs_gae}\\[0.3em]
  \hyperref[sec:ext_analysis]{B}\quad \textbf{Extended analysis}\dotfill\pageref{sec:ext_analysis}\\[0.3em]
  \hyperref[sec:setup_svg_sm]{C}\quad \textbf{Image-to-SVG appendix}\dotfill\pageref{sec:setup_svg_sm}\\
  \hspace*{1em}\hyperref[sec:test_time_extended]{C.1}\quad Test-time scaling\dotfill\pageref{sec:test_time_extended}\\
  \hspace*{1em}\hyperref[sec:svg_closure]{C.2}\quad Rendering boundaries\dotfill\pageref{sec:svg_closure}\\
  \hspace*{1em}\hyperref[sec:svg_train_params]{C.3}\quad Training parameters\dotfill\pageref{sec:svg_train_params}\\
  \hspace*{1em}\hyperref[sec:svg_model_scale]{C.4}\quad Impact of Base Model Size\dotfill\pageref{sec:svg_model_scale}\\
  \hspace*{1em}\hyperref[sec:svg_baselines]{C.5}\quad Baseline parameters\dotfill\pageref{sec:svg_baselines}\\
  \hspace*{1em}\hyperref[sec:svg_metrics]{C.6}\quad Metrics\dotfill\pageref{sec:svg_metrics}\\
  \hspace*{1em}\hyperref[sec:svg_budget]{C.7}\quad Generation budget and Length Penalty\dotfill\pageref{sec:svg_budget}\\[0.3em]
  \hyperref[sec:setup_tikz_sm]{D}\quad \textbf{Image-to-TikZ appendix}\dotfill\pageref{sec:setup_tikz_sm}\\
  \hspace*{1em}\hyperref[sec:tikz_ablation]{D.1}\quad Reward ablation\dotfill\pageref{sec:tikz_ablation}\\
  \hspace*{1em}\hyperref[sec:tikz_closure]{D.2}\quad Rendering boundaries\dotfill\pageref{sec:tikz_closure}\\
  \hspace*{1em}\hyperref[sec:tikz_train_params]{D.3}\quad Training parameters\dotfill\pageref{sec:tikz_train_params}\\
  \hspace*{1em}\hyperref[sec:tikz_baselines]{D.4}\quad Baseline parameters\dotfill\pageref{sec:tikz_baselines}\\
  \hspace*{1em}\hyperref[sec:tikz_metrics]{D.5}\quad Metrics\dotfill\pageref{sec:tikz_metrics}\\
\endgroup

\section{Connection to Potential-Based Reward Shaping and GAE}
\label{sec:pbrs_gae}

In this section, we interpret our method through potential-based reward shaping (PBRS)~\citep{ng1999policy} and Generalized Advantage Estimation (GAE)~\citep{schulman2015high}.
PBRS provides conditions for modifying the reward function of a Markov decision process (MDP)
that preserve the optimal policy while potentially reducing learning time. A shaped reward becomes
\begin{equation}
r_t' = r_t + \gamma\Phi(s_t) - \Phi(s_{t-1}),
\end{equation}
where $r_t$ is the unmodified reward of the original MDP for the transition $s_{t-1} \xrightarrow{a_t} s_t$, $\gamma$ is its discount factor, and $\Phi$ is an arbitrary function of the state that does not depend on the policy. 
With $\Phi(s_T)=0$ assumed at a terminal state, the shaping terms telescope in the discounted return of a trajectory, $\sum_{t=1}^{T}\gamma^{t-1} r'_t = \sum_{t=1}^{T}\gamma^{t-1} r_t - \Phi(s_0)$, shifting it by a policy-independent constant and leaving the optimal policy unchanged.

For LLMs, generation is commonly modeled as a finite-horizon, undiscounted MDP ($\gamma=1$, omitted from here on for simplicity), with a nonzero reward only at the terminal state ($r_T = R$, $r_{t\neq T}=0$)~\citep{williams1992simple, deepseek-math}. The state $s_{t-1}=(x,y_{<t})$ denotes a prefix of tokens generated so far, the action $a_t = y_t$ is the next generated token, and the LLM is a policy $\pi_\theta(y_t \mid x, y_{<t})$.
In this LLM setting, the shaped reward can be intuitively viewed as a composition of an outcome-driven reward and a process reward: 
\begin{equation}
r'_t = \underbrace{\mathbbm{1}[t=T]\,R}_{r^{\mathrm{Outcome}}_t} + \underbrace{\Phi(s_t)-\Phi(s_{t-1})}_{r^{\mathrm{Process}}_t}.
\end{equation}

The policy gradient weights each token's log-probability by its advantage, which corresponds to the return from that token minus a baseline~\citep{williams1992simple, schulman2015high} and must be estimated in practice since the ideal baseline is unknown. With the shaped reward, the return from token $t$ decomposes as
\begin{equation}
\sum_{k=t}^{T} r'_k = \underbrace{R}_{\text{Outcome}} + \underbrace{\sum_{k=t}^{T}\big(\Phi(s_k)-\Phi(s_{k-1})\big)}_{\text{Process}}.
\label{eq:rtg_decomposition}
\end{equation}
The same decomposition applies to advantages, yielding the combined advantage $A_{i,t} = A_i^{\mathrm{Outcome}} +  A_{i,t}^{\mathrm{Process}}$ (Eq.~6), and allowing us to estimate each term separately.
For $A_i^{\mathrm{Outcome}}$, we use the standard GRPO estimator based on group statistics \cite{deepseek-math}, and $A_{i,t}^{\mathrm{Process}}$ is obtained below.

We first define $r^{\mathrm{Process}}_t$ through our render-progress reward $\Delta_j$ (Eq.~\ref{eq:delta}), which can be viewed as a difference of potentials with $\Phi(s_t) = F_{j(t)}$, where $F$ is the visual score up to the last rendered segment $j$ (Eq.~\ref{eq:vis_score}). Thus, $\Phi$ is constant $r^{\mathrm{Process}}_t=\Phi(s_t)-\Phi(s_{t-1})=0$  within segments.
For the corresponding $A^{\mathrm{Process}}_{i,t}$, we follow GAE estimator,
which reduces the variance of policy gradient estimates at the cost of a tolerable level of bias by exponentially weighting future TD residuals $\delta_t = r_t + \gamma V(s_t) - V(s_{t-1})$ with a parameter $\lambda$: 
$A^{\mathrm{GAE}}_t = \sum_{k\ge t}(\gamma\lambda)^{k-t}\,\delta_k$.
Our $r^{\mathrm{Process}}_t$ can be viewed as a special case of $\delta_t$ with $V=\Phi$ and $\gamma=1$.
Thus, GAE yields our Eq.~\ref{eq:progress_credit}:
\begin{equation}
\scalebox{0.9}{$\displaystyle
A^{\mathrm{GAE}}_t
= \sum_{k\ge t}(\gamma\lambda)^{k-t}\, r^{\mathrm{Process}}_k
= \sum_{j:\, b_j\ge t}\lambda^{\,b_j-t}\,\Delta_j
= A^{\mathrm{Process}}_t.
$}
\label{eq:gae_process}
\end{equation}
Scaling the potential to $\Phi = \alpha F$ effectively scales the process advantage, allowing to control its strength, as in Eq.~6.

For $\lambda=1$, the sum telescopes to $A^{\mathrm{Process}}_t = \Phi(s_T) - \Phi(s_{t-1}) = R - F_{j(t-1)}$, (as $\Phi(s_T)=F_M = R$). 
Together with the outcome term, this yields a total shaped return from token $t$ of $2R-F_{j(t-1)}$. 
Since $F_{j(t-1)}$ serves as a baseline independent of token $y_t$, the process term optimizes the same objective as $A^{\mathrm{Outcome}}$, so the optimal policy is preserved as in PBRS, despite $\Phi(s_T)\neq0$.

The case of $\lambda < 1$ introduces bias into the gradient estimate, since intermediate potentials no longer cancel, breaking the policy invariance guarantee and weighting nearer render events higher than distant ones. \citet{schulman2015high} report optimal values of $\lambda \in [0.9, 0.99]$ on their control tasks, reflecting the bias--variance tradeoff. We observe similar optima: $\lambda = 0.9$ for Image-to-SVG and $\lambda = 0.95$ for Image-to-TikZ (Figs.~\ref{fig:analysis}b,~\ref{fig:analysis_tikz}b). 

Notably, neither PBRS nor GAE requires the function $\Phi$ to be monotonic between segments, potentially making our method flexible to choices of the visual score function for new tasks.

\vspace{5pt}

\section{Extended analysis}
\label{sec:ext_analysis}
\paragraph{Hyperparameters analysis.}
\label{sec:hyperparams}

\begin{figure*}[t!]
  \centering
  \includegraphics[width=\linewidth]{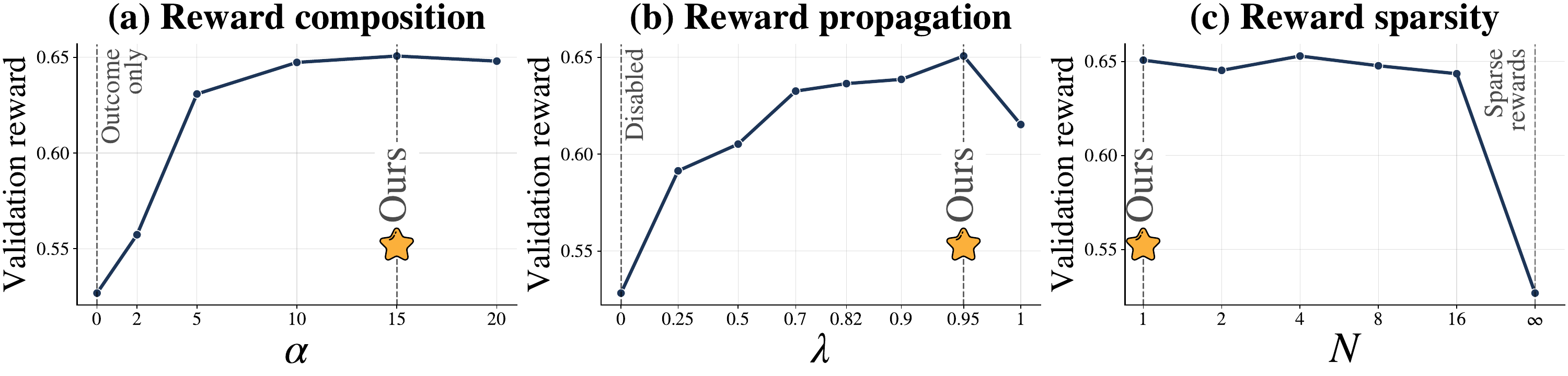}
  \caption{ \textbf{Ablations of process supervision on DaTikZ-v3 validation set.} (a) Increasing the process contribution improves performance and plateaus around $\alpha\in[10,15]$. (b) Reward propagation performs best at $\lambda=0.95$. (c) Performance is robust to render sparsity up to $N=8$ and degrades at higher sparsity. Stars mark our default settings. 
  }
  \label{fig:analysis_tikz}
\end{figure*}

\begin{figure*}[t!]
  \centering
  \begin{minipage}[t]{0.49\linewidth}
    \centering
    {\small \textbf{(a) Image-to-SVG}}
    \includegraphics[width=\linewidth]{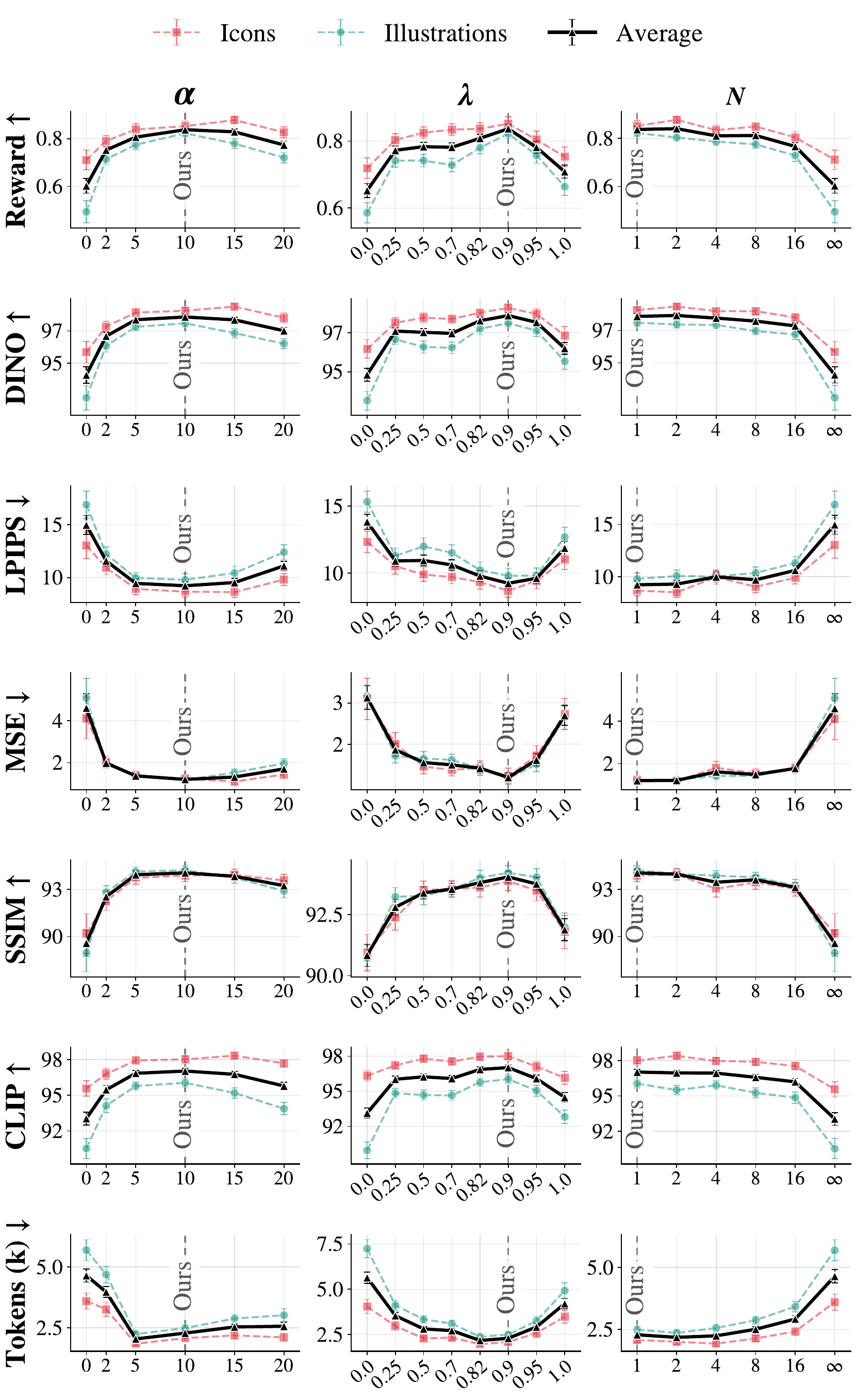}
  \end{minipage}\hfill
  \begin{minipage}[t]{0.49\linewidth}
    \centering
    {\small \textbf{(b) Image-to-TikZ}}
    \includegraphics[width=\linewidth]{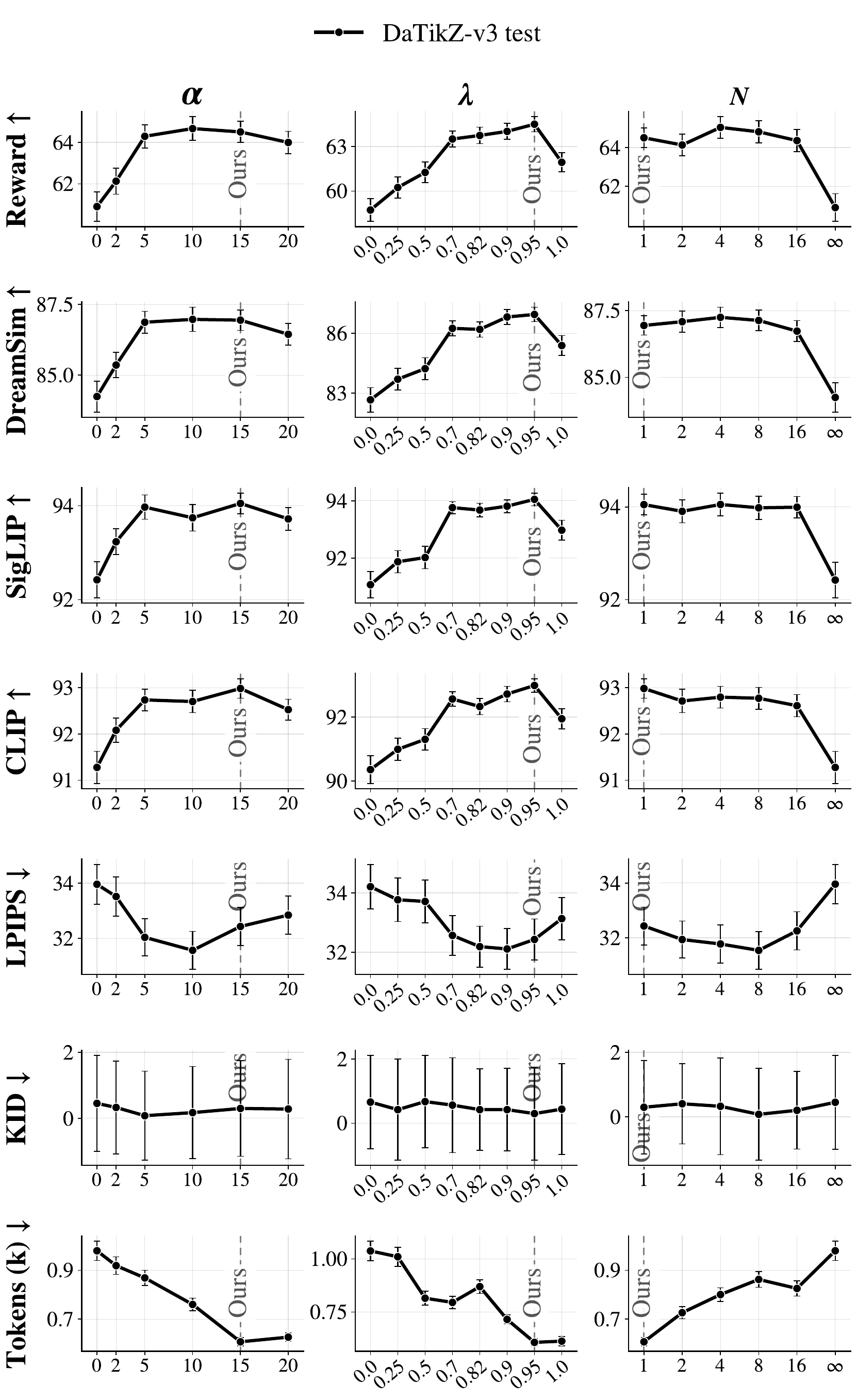}
  \end{minipage}

  \caption{\textbf{Process supervision parameters on test sets.}
   (a) Image-to-SVG on MMSVGBench.
(b) Image-to-TikZ on DaTikZ v3.
Settings selected on each task's validation set (dashed lines,
\emph{Ours}) achieve the highest test reward on SVG and
near-optimal reward on TikZ. Error bars show S.E.M., except for KID, where they show the
standard deviation across estimates, each using a
50-image subset.
   }
  \label{fig:analysis_extended}
\end{figure*}
  
Fig.~\ref{fig:analysis_tikz} presents the same analysis as
Sec.~\ref{sec:analysis} for Image-to-TikZ, leading to the selected
configuration of $\alpha=15$, $\lambda=0.95$, $N=1$.
Fig.~\ref{fig:analysis_extended} reports test-set metrics for
the same ablations for both tasks.
The Image-to-SVG parameters ($\alpha=10$, $\lambda=0.9$, $N=1$)
remain competitive for Image-to-TikZ, and vice versa.
Task-specific tuning thus adds only modest gains, supporting
either configuration as a reasonable default across tasks. Interestingly, Image-to-TikZ is less sensitive to rendering
frequency than Image-to-SVG, with reconstruction quality
remaining largely stable as $N$ increases from $1$ to $16$.

\paragraph{Computational cost.}

\begin{table*}[t]
  \centering
  \resizebox{\textwidth}{!}{
  \begin{tabular}{lccccccc|ccccccc}
              \toprule
              & \multicolumn{7}{c}{\textbf{MMSVGBench-Illustrations}} & \multicolumn{7}{c}{\textbf{MMSVGBench-Icons}} \\
              \cmidrule(lr){2-8} \cmidrule(lr){9-15}
                                          & DINO $\uparrow$ & LPIPS $\downarrow$ & MSE $\downarrow$ & SSIM $\uparrow$ & CLIP $\uparrow$ &
      Aesthetic $\uparrow$ & Tokens $\downarrow$ & DINO $\uparrow$ & LPIPS $\downarrow$ & MSE $\downarrow$ & SSIM $\uparrow$ & CLIP $\uparrow$ &
      Aesthetic $\uparrow$ & Tokens $\downarrow$ \\
              \midrule
              + Outcome                    & 92.85 & 16.89 & 5.08 & 88.97 & 90.52 & 4.88 & 5.7k & 95.68 & 13.04 & 4.12 & 90.20 & 95.57 & 4.90 & 3.6k \\
              + Ours (time-aligned)        & \underline{97.22} & \underline{9.97} & \underline{1.22} & \textbf{94.24} & \underline{95.59} & \underline{5.05} & \textbf{2.4k} & \underline{98.11} & \underline{9.00} & \textbf{1.21} & \underline{93.81} & \underline{97.87} & \textbf{5.01} & \textbf{2.0k} \\
              + Ours                       & \textbf{97.48} & \textbf{9.79} & \textbf{1.17} & \underline{94.22} & \textbf{96.05} & \textbf{5.09} & \underline{2.5k} & \textbf{98.26} & \textbf{8.67} & \underline{1.23} & \textbf{93.88} & \textbf{98.01} & \underline{4.98} & \underline{2.1k} \\
              \bottomrule
          \end{tabular}

  \vspace{1em}
  }

    \resizebox{0.6\textwidth}{!}{
    \begin{tabular}{lcccccc}
          \toprule
          & DreamSim $\uparrow$ & SigLIP $\uparrow$ & CLIP $\uparrow$ & LPIPS $\downarrow$ & KID $\downarrow$ & Tokens $\downarrow$ \\
          \midrule
          + Outcome                      & 84.2{\scriptsize$\pm$0.5} & 92.4{\scriptsize$\pm$0.4} & 91.3{\scriptsize$\pm$0.3} & 34.0{\scriptsize$\pm$0.7} & \underline{0.45}{\scriptsize$\pm$1.45} & 1.0k \\
          + Ours (time-aligned)          & \underline{85.9}{\scriptsize$\pm$0.4} & \underline{93.5}{\scriptsize$\pm$0.2} & \underline{92.3}{\scriptsize$\pm$0.2} & \underline{33.7}{\scriptsize$\pm$0.7} & 0.56{\scriptsize$\pm$1.35} & \underline{0.7k} \\
          + Ours                         & \textbf{86.9}{\scriptsize$\pm$0.4} & \textbf{94.1}{\scriptsize$\pm$0.2} & \textbf{93.0}{\scriptsize$\pm$0.2} & \textbf{32.4}{\scriptsize$\pm$0.7} & \textbf{0.30}{\scriptsize$\pm$1.45} & \textbf{0.6k} \\
          \bottomrule
      \end{tabular}
      }
  
  \caption{\textbf{Quantitative evaluation with time-aligned checkpoints.} (top) Image-to-SVG on MMSVGBench, (bottom) Image-to-TikZ on DaTikZ-v3 test set. 
At the training time of outcome-only RL, our method uses
44\% (SVG) and 54\% (TikZ) of its full post-training steps,
yet improves all per-image quality metrics and produces shorter programs.
Reported $\pm$ values denote S.E.M. for per-image metrics.
For KID, they denote standard deviation across estimates from
50-image subsets, reflecting sensitivity to subset sampling.
  }
  \label{tab:metrics_runtime}
\end{table*}

\begin{figure}[t]
  \centering
  \includegraphics[width=1.0\linewidth, keepaspectratio]{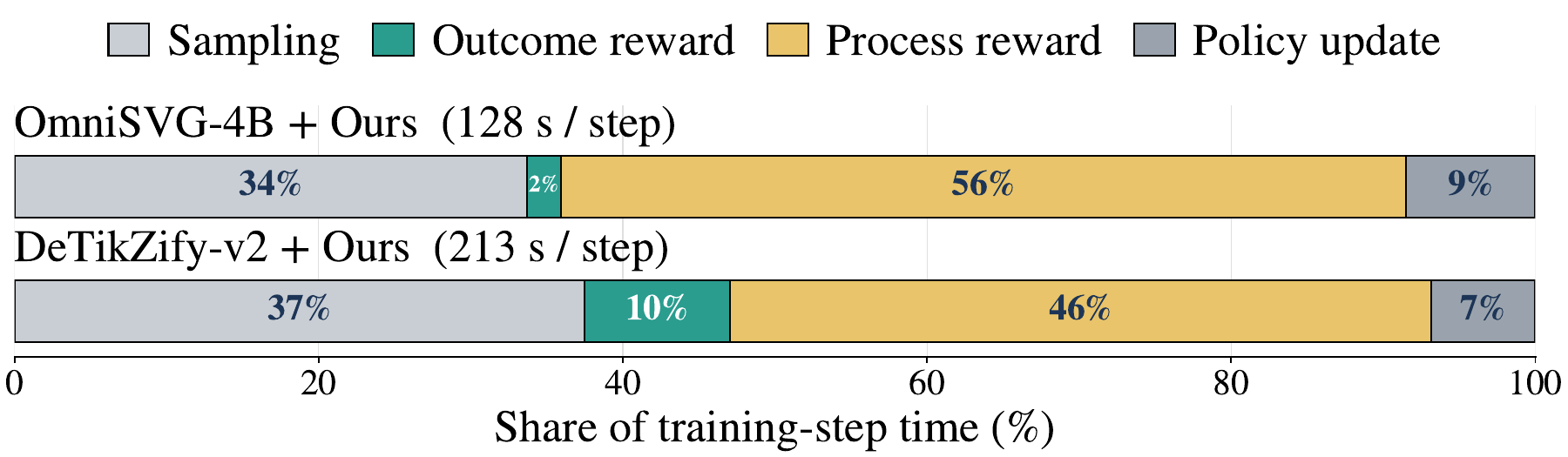}
    \caption{\textbf{Training runtime breakdown (\% and seconds).}
      Average step wall-clock time \% across phases: outcome and process rewards cover CPU rendering and visual scoring. Policy update includes forward, backward, and optimizer step. }
    \label{fig:runtime}
\end{figure}
  
In Fig.~\ref{fig:runtime}, we report an average wall-clock time of training step stages on a single NVIDIA B200 GPU with 20 cores of an Intel Xeon Platinum 8570 CPU, excluding evaluation and checkpointing, for a batch of 128 completions. The outcome reward renders and scores each sample once, while the process reward additionally processes intermediate prefixes at command boundaries, taking roughly half of a step runtime in the Image-to-SVG task and Image-to-TikZ tasks. 
Complementing the training dynamics in Fig.~\ref{fig:converge}, Tab.~\ref{tab:metrics_runtime} compares outcome-only training with our method evaluated at a time-matched checkpoint corresponding to roughly half of the training, where our method still outperforms across almost all metrics. This further shows that the extra per-step cost is justified.

\section{Image-to-SVG setup details}
\label{sec:setup_svg_sm}

\subsection{Test-time scaling}
\label{sec:test_time_extended}
\begin{figure*}[h]
  \centering
  \includegraphics[width=1.0\linewidth, keepaspectratio]{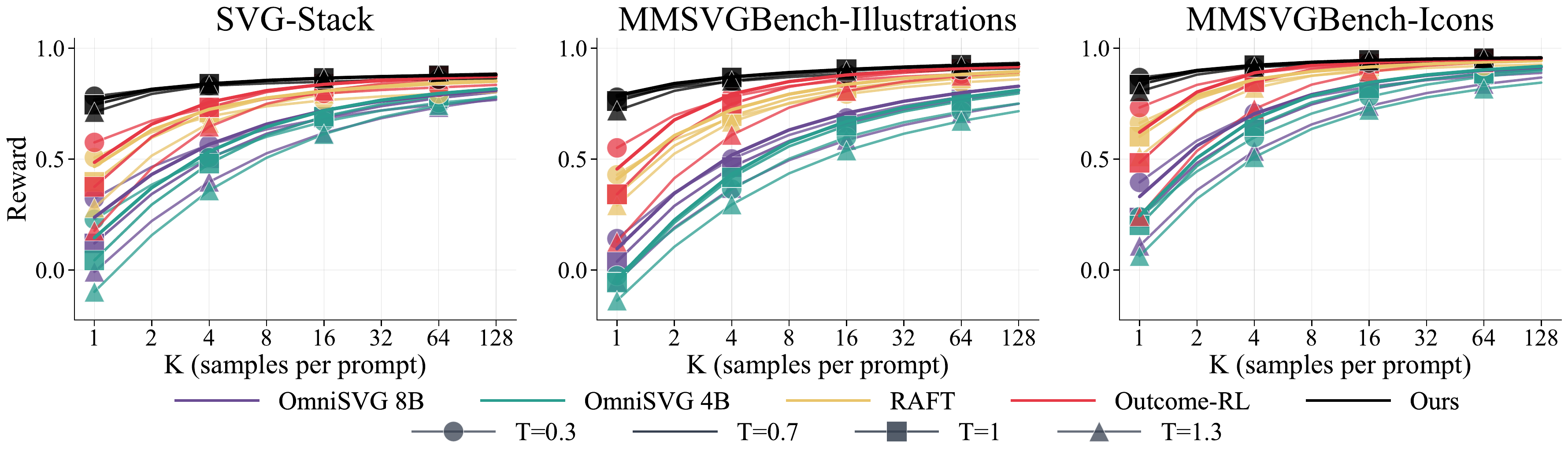}
    \caption{
    \textbf{Best-of-$K$ sampling on SVG generation.}
    For each image, we sample $K$ candidate SVGs and report the mean reward of the best candidate. Across multiple datasets and temperature values, our method performs best across all sampling budgets, with the largest gains at small $K$.}
    \label{fig:best_of_k_extended}
\end{figure*}

We extend the test-time scaling analysis from Sec.~\ref{sec:analysis} across three evaluation sets and four sampling temperatures, $T\in\{0.3,0.7,1.0,1.3\}$ in Fig.~\ref{fig:best_of_k_extended}.
Our method achieves the highest reward across nearly all sampling budgets and temperatures, with the largest advantage at small $K$. As $K$ increases, the gap to the baselines narrows, consistent with our method assigning greater probability to high-quality generations rather than relying on large sampling budgets to discover them. This trend is consistent across SVG-Stack, MMSVGBench-Illustrations, and MMSVGBench-Icons. 

\subsection{Rendering boundaries}
\label{sec:svg_closure}

\begin{figure*}[t]
  \centering
  \includegraphics[width=1.0\linewidth, keepaspectratio]{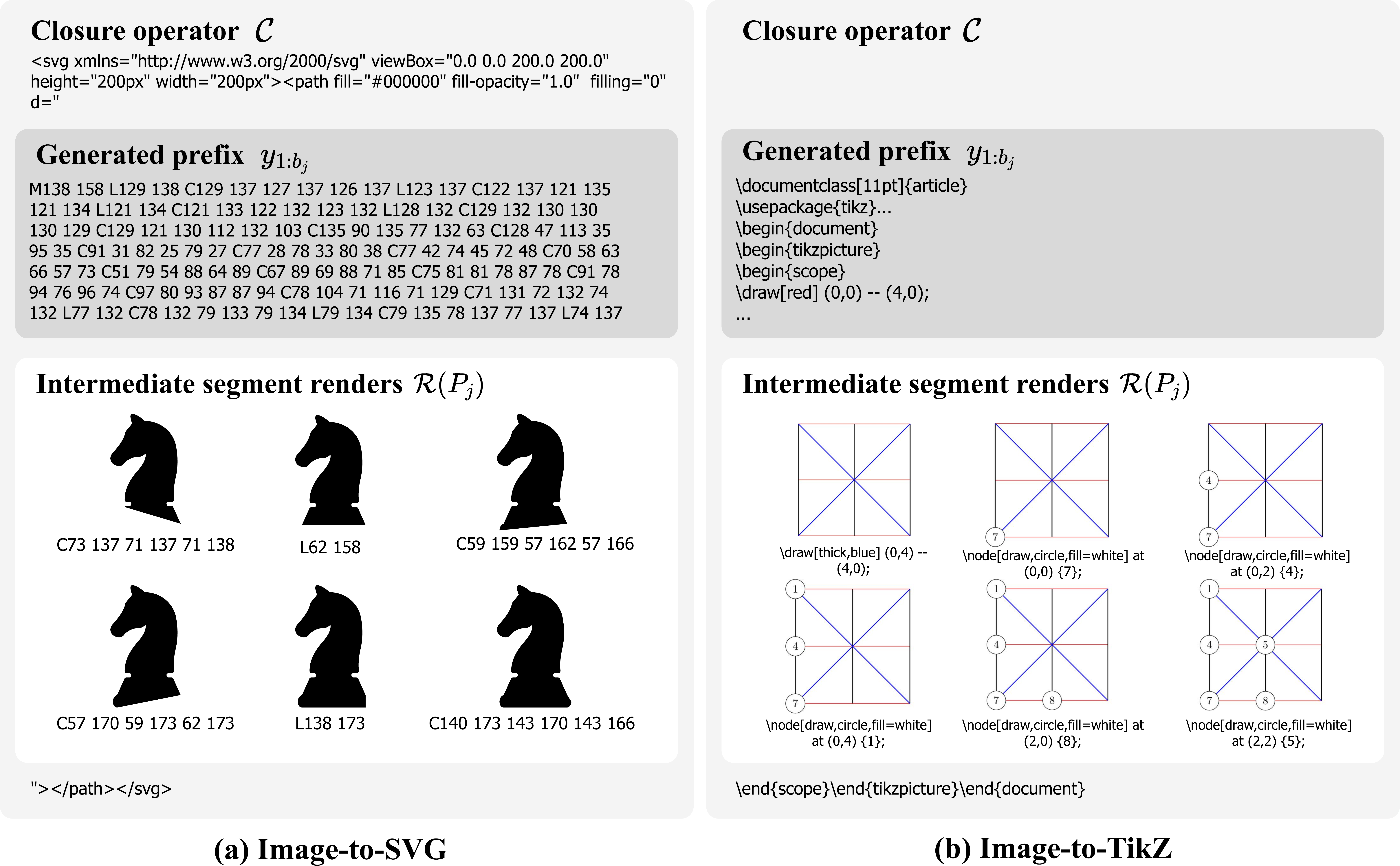}
    \caption{
    \textbf{Prefix closure} for (a) Image-to-SVG, (b) Image-to-TikZ. A model generates a program code $y$ that is separated into granular segments $y_{1:b_j}$, well-formed by a closure operator $\mathcal{C}$ into $P_j=\mathcal{C}(y_{1:b_j})$
 and rendered. This process is defined to align with the task and a tokenizer. Specifically, OmniSVG does not generate the \texttt{<svg>} tag, while in DeTikZify the opening code is part of the sequence. 
    }
    \label{fig:closure}
\end{figure*}

After each generated token, we check whether the current SVG drawing
command (line, curve, arc, or color) has all its required parameters.
This check is inexpensive because the command type determines which
parameters to expect. If the command is incomplete, we assign the token
zero progress reward. In our default setting ($N=1$), every
token that completes a command marks a rendering boundary $b_j$.
Roughly 15\% of generated tokens mark such boundaries.

At each boundary, the closure operator $\mathcal{C}$ wraps the entire
prefix $y_{1:b_j}$ in the \texttt{<svg>} and \texttt{<path>} markup
needed to turn OmniSVG's drawing commands into a standalone SVG
(Fig.~\ref{fig:closure}a). We can then render the drawing through the
latest completed segment, even if the model later extends the same
path. For example, when filling a path, the renderer connects the
current endpoint back to the start of the subpath. We score the
resulting image and assign the boundary token the change in visual
score since the preceding render, $\Delta_j = F_j - F_{j-1}$, as its
render-progress reward.

\subsection{Training parameters}
\label{sec:svg_train_params}

\begin{table*}[t]
  \centering
  \setlength{\tabcolsep}{3pt}
  \resizebox{\textwidth}{!}{%
  \begin{tabular}{@{}lccccccc|ccccccc@{}}
      \toprule
      & \multicolumn{7}{c}{\textbf{MMSVGBench-Illustrations}}
      & \multicolumn{7}{c}{\textbf{MMSVGBench-Icons}} \\
      \cmidrule(lr){2-8} \cmidrule(lr){9-15}
      & DINO $\uparrow$ & LPIPS $\downarrow$ & MSE $\downarrow$
      & SSIM $\uparrow$ & CLIP $\uparrow$ & Aesthetic $\uparrow$
      & Tokens $\downarrow$
      & DINO $\uparrow$ & LPIPS $\downarrow$ & MSE $\downarrow$
      & SSIM $\uparrow$ & CLIP $\uparrow$ & Aesthetic $\uparrow$
      & Tokens $\downarrow$ \\
      \midrule
      700 samples
      & \textbf{97.48} & \textbf{9.79} & \textbf{1.17}
      & \textbf{94.22} & \textbf{96.05} & \textbf{5.09}
      & \underline{2.5k}
      & \underline{98.26} & \underline{8.67} & \underline{1.23}
      & \underline{93.88} & \underline{98.01} & \underline{4.98}
      & \underline{2.1k} \\
      10k samples
      & \underline{97.26} & \underline{10.11} & \underline{1.27}
      & \underline{94.11} & \underline{95.35} & \underline{5.08}
      & \textbf{2.2k}
      & \textbf{98.49} & \textbf{8.42} & \textbf{1.20}
      & \textbf{94.12} & \textbf{98.35} & \textbf{5.02}
      & \textbf{1.9k} \\
      \bottomrule
  \end{tabular}%
  }
  \caption{\textbf{Effect of the number of training samples.}
  Increasing the training set from 700 to 10k samples yields
  marginal gains on Icons and mild degradation on Illustrations,
  while producing shorter programs on both subsets.
  We therefore use the smaller training set in all other experiments.}
  \label{tab:data_scaling}
\end{table*}

The visual similarity function $S$ between SVG renders and a target is defined following prior work \citep{rodriguez2023starvector} as a scale-invariant normalized L2 that emphasizes structural and color discrepancies:
\begin{equation}
S = \text{clip}\!\left(1 - \frac{1}{|I|} \left\| I_{\text{in}}^{\text{norm}} - I_{\text{pred}}^{\text{norm}} \right\|_2^2,\ -1,\ 1\right),
\label{eq:img_reward}
\end{equation}
where $I^{\text{norm}}$ is z-score normalized image. $S$ lies within $[-1, 1]$, and for blank or malformed SVGs it is equal to $-1$.

We fine-tune OmniSVG-4B with AdamW~\citep{loshchilov2019decoupled} and a constant learning rate scheduler, using a LoRA adapter~\citep{hu2022lora} ($r{=}64$, $\alpha_{\text{LoRA}}{=}128$, dropout $0.05$) on the attention (query, key, value, output) and MLP (gate, up, down) projections of the LLM backbone. 
Each optimizer step uses 8 unique images with a group size of 16, resulting in 128 rollouts sampled at temperature $1.1$. 
We tune the learning rate over $\text{lr} \in \{5\text{e-}5, 1\text{e-}4\}$ and the KL coefficient over $\beta\in\{0, 0.01, 0.02\}$, selecting 
$(\text{lr}, \beta)=(5\text{e-}5, 0)$.
Method-specific hyperparameters ($\alpha$, $\lambda$, $N$) are ablated one at a time using the selected configuration (Fig.~6). Each run is trained for 400 steps on a single NVIDIA B200 GPU.
The training set size is 700 samples of svg-stack~\citep{rodriguez2023starvector}.
Tab.~\ref{tab:data_scaling} shows that scaling to 10k samples (a $14\times$ increase) brings no consistent gain in reconstruction quality: 700 samples wins on Illustrations, 10k on Icons, with marginal difference in both cases. This aligns with recent findings that RL fine-tuning can saturate on remarkably few examples~\citep{wang2025reinforcement}, and we adopt the smaller, cheaper set.

\subsection{Impact of Base Model Size}
\label{sec:svg_model_scale}

\setlength{\tabcolsep}{2pt}
      \begin{table*}[t]
          \centering
          \resizebox{\linewidth}{!}{
          \begin{tabular}{lccccccc|ccccccc}
              \toprule
              & \multicolumn{7}{c}{\textbf{MMSVGBench-Illustrations}} & \multicolumn{7}{c}{\textbf{MMSVGBench-Icons}} \\
              \cmidrule(lr){2-8} \cmidrule(lr){9-15}
                                          & DINO $\uparrow$ & LPIPS $\downarrow$ & MSE $\downarrow$ & SSIM $\uparrow$ & CLIP $\uparrow$ &
      Aesthetic $\uparrow$ & Tokens $\downarrow$ & DINO $\uparrow$ & LPIPS $\downarrow$ & MSE $\downarrow$ & SSIM $\uparrow$ & CLIP $\uparrow$ &
      Aesthetic $\uparrow$ & Tokens $\downarrow$ \\
              \midrule
              OmniSVG-4B                   & 85.48 & 22.26 & 5.11 & 89.70 & 82.03 & 4.55 & 11.3k & 89.21 & 19.82 & 5.80 & 87.96 & 88.42 & 4.65 & 8.4k \\
              OmniSVG-8B                   & 88.81 & 21.15 & 4.98 & 88.85 & 85.57 & \underline{4.68} & 9.1k & 92.04 & 18.09 & 4.92 & 89.28 & 91.89 & 4.79 & 6.0k \\
              OmniSVG-4B + Ours    & \textbf{97.48} & \underline{9.79} & \textbf{1.17} & \underline{94.22} & \underline{96.05} & \textbf{5.09} & \textbf{2.5k} & \underline{98.26} & \underline{8.67} & \underline{1.23} & \underline{93.88} & \underline{98.01} & \underline{4.98} & \textbf{2.1k} \\
              OmniSVG-8B + Ours    & \underline{97.34} & \textbf{9.56} & \underline{1.20} & \textbf{94.32} & \textbf{96.29} & \textbf{5.09} & \underline{3.0k} & \textbf{98.38} & \textbf{8.47} & \textbf{1.12} & \textbf{94.25} & \textbf{98.47} & \textbf{4.99} & \underline{2.3k} \\
              \bottomrule
          \end{tabular}
          }
          \caption{\textbf{Effect of IR4RL at 4B and 8B scale.} Despite a significant margin between 4B and 8B base models, applying our method to both models produces similar results, motivating us to adopt a smaller variant.}
          \label{tab:model_scaling}
      \end{table*}

Table~\ref{tab:model_scaling} compares our method applied to OmniSVG-4B and OmniSVG-8B.
The 8B model follows the same training recipe as the
4B model, except that we scale the learning rate by $0.7\times$ to account for model size, following~\citep{yang2021neurips-tuning}, which we found to work best for 8B.
Before post-training, the 8B SFT model is stronger under greedy decoding, but
Fig.~\ref{fig:best_of_k} shows that the 4B and 8B models approach similar
performance at larger sampling budgets. After post-training, this gap largely
disappears, while the 8B model produces approximately $20\%$ longer
completions. This is consistent with prior observations that RL can increase
the probability of high-quality generations already reachable under the base policy~\citep{deepseek-math,yue2025limit}. We therefore use OmniSVG-4B for the
remaining experiments, as it achieves comparable final results at lower computational cost.

\subsection{Baseline Parameters}
\label{sec:svg_baselines}

DiffVG was run with 500 iterations and 128 paths, and LIVE with 200 iterations and 16 paths, following their default settings. General-purpose (M)LLMs (Qwen3-VL-235B~\citep{Qwen3-VL}, Gemini~3~Flash~\citep{gemini3flash}, Sonnet~5~\citep{claudesonnet5}, GPT-5.2~\citep{gpt52}) are prompted with the input image via each provider's API using the system prompt in
Fig.~\ref{fig:sys_prompts}a. StarVector-8B~\citep{rodriguez2023starvector} and
InternSVG-8B~\citep{wang2025internsvg} are decoded with each model's default sampling parameters. OmniSVG's base models (4B, 8B) are sampled with default top-$p=0.95$, top-$k=50$, temperature $0.3$, and all the discussed post-training variants follow the same sampling configuration. 
RAFT~\citep{dong2023raft} is trained for 15 iterations, sampling 32 completions per input per iteration, selecting the highest-reward sample, and fine-tuning for two SFT epochs on the selected set. 

\subsection{Metrics}
\label{sec:svg_metrics}
CLIP is computed image-vs-image rather than image-vs-text. Both CLIP and DINO use their base-size checkpoints (ViT-B/32, dinov2-base), and LPIPS uses the AlexNet backbone. Aesthetic is a no-reference score of the generated image alone. 
For fair comparison, token counts for all models use Qwen2.5-VL-3B-Instruct tokenizer\cite{bai2025qwen25vl}, applied to the extracted SVG markup, following OmniSVG evaluation setup. For the API responses, the leaked prose and thinking-trace text is additionally taken into account.

\begin{figure}[t]
  \centering
  \includegraphics[width=0.8\linewidth, keepaspectratio]{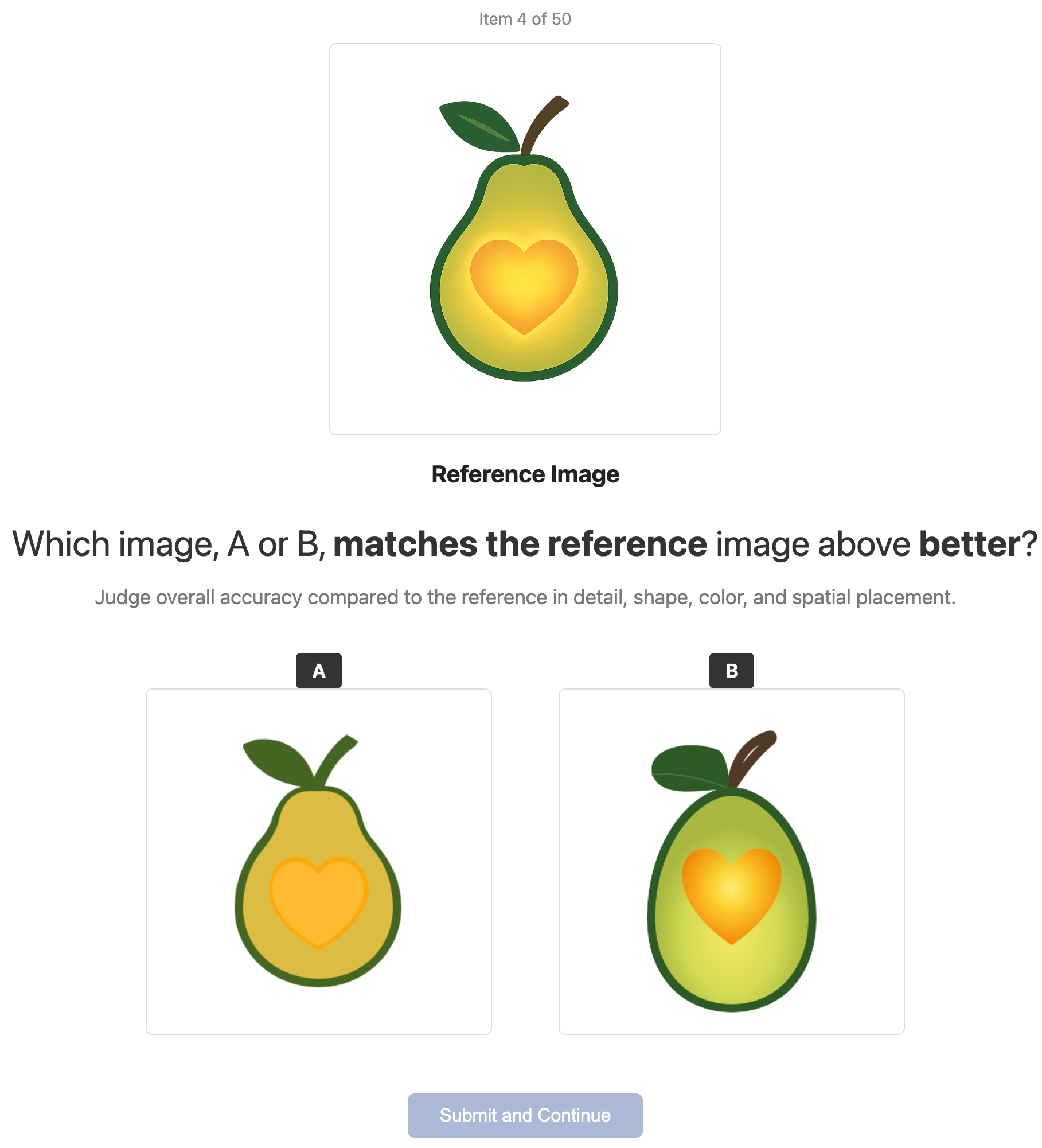}
    \caption{
    \textbf{User study interface.} Example of an interface with a question comparing Ours to Gemini 3 Flash.
    }
    \label{fig:user_study_ux}
\end{figure}

\begin{figure*}[t]
  \centering
  \includegraphics[width=1.0\linewidth, keepaspectratio]{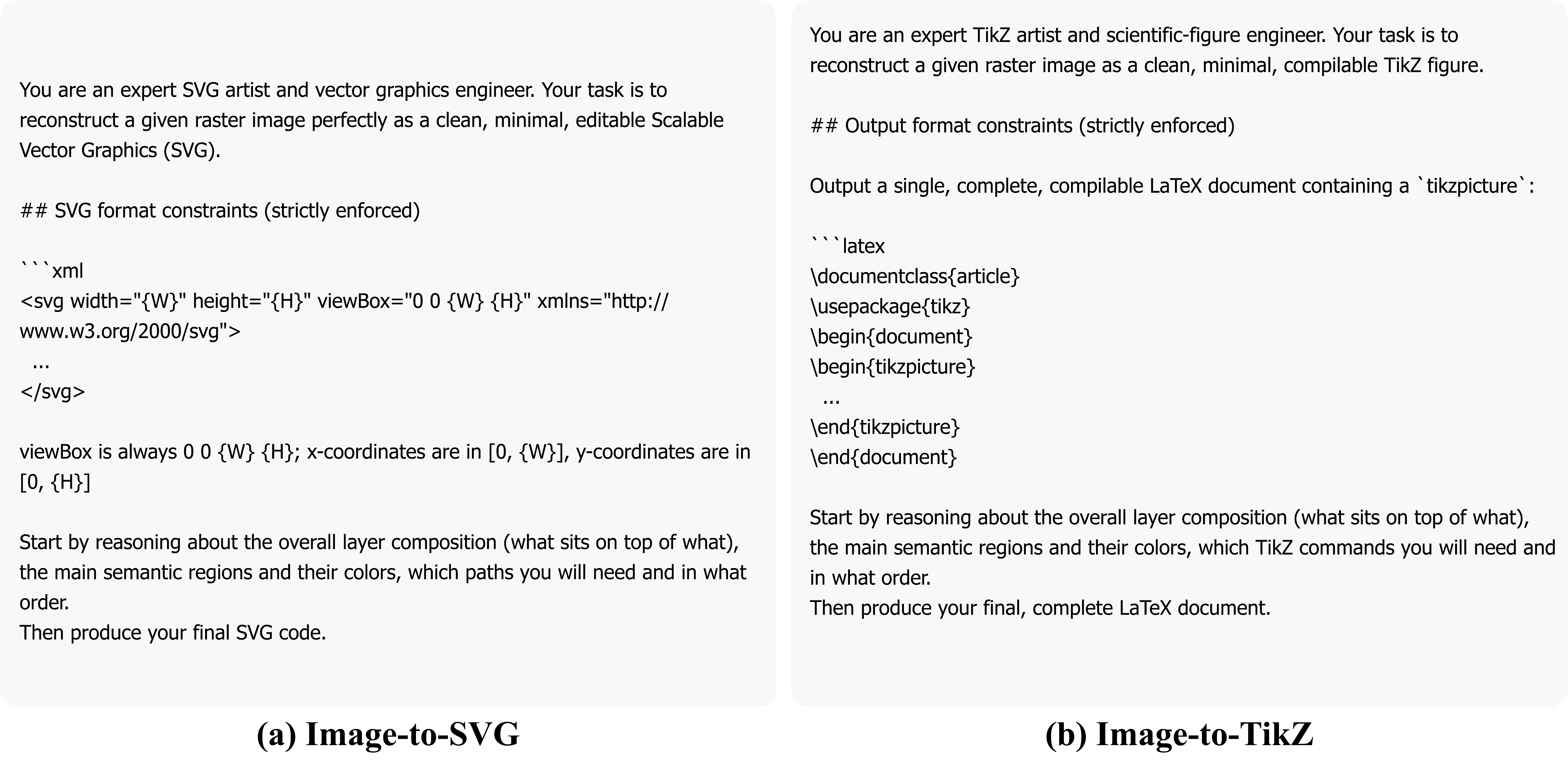}
    \caption{
    \textbf{API evaluation prompts.} System prompts used for evaluation with closed-source API models of (a) Image-to-SVG, (b) Image-to-TikZ.
    }
    \label{fig:sys_prompts}
    \vspace{-10pt}
\end{figure*}

A user study was conducted on Prolific~\citep{prolific2024} platform with 50 samples drawn at random from MMSVGBench. For each baseline, participants saw the reference image alongside two candidates, ours and the baseline's, presented in random order as a two-alternative forced-choice (interface in Fig.~\ref{fig:user_study_ux}). They were asked: \textit{"Which image, A or B, matches the reference image above better? Judge overall accuracy compared to the reference in detail, shape, color, and spatial placement."}. Each comparison was rated by 6 participants, giving 1500 judgments from 30 participants in total. Participants provided informed consent before participation and were compensated through Prolific. The study collected only preference judgments and did not involve sensitive personal data. 

As an automatic counterpart with a VLM-as-a-judge, we ran the same comparisons through Gemini~3~Flash~\citep{gemini3flash}, providing it with the input and the question matching the one seen by humans, and asking it to answer with A or B plus a brief justification.

\subsection{Inference Budget and Length Penalty}
\label{sec:svg_budget}

\begin{figure*}[t!]
  \centering
  \begin{minipage}[t]{0.49\linewidth}
    \vspace{0pt}
    \centering
    \includegraphics[width=\linewidth]{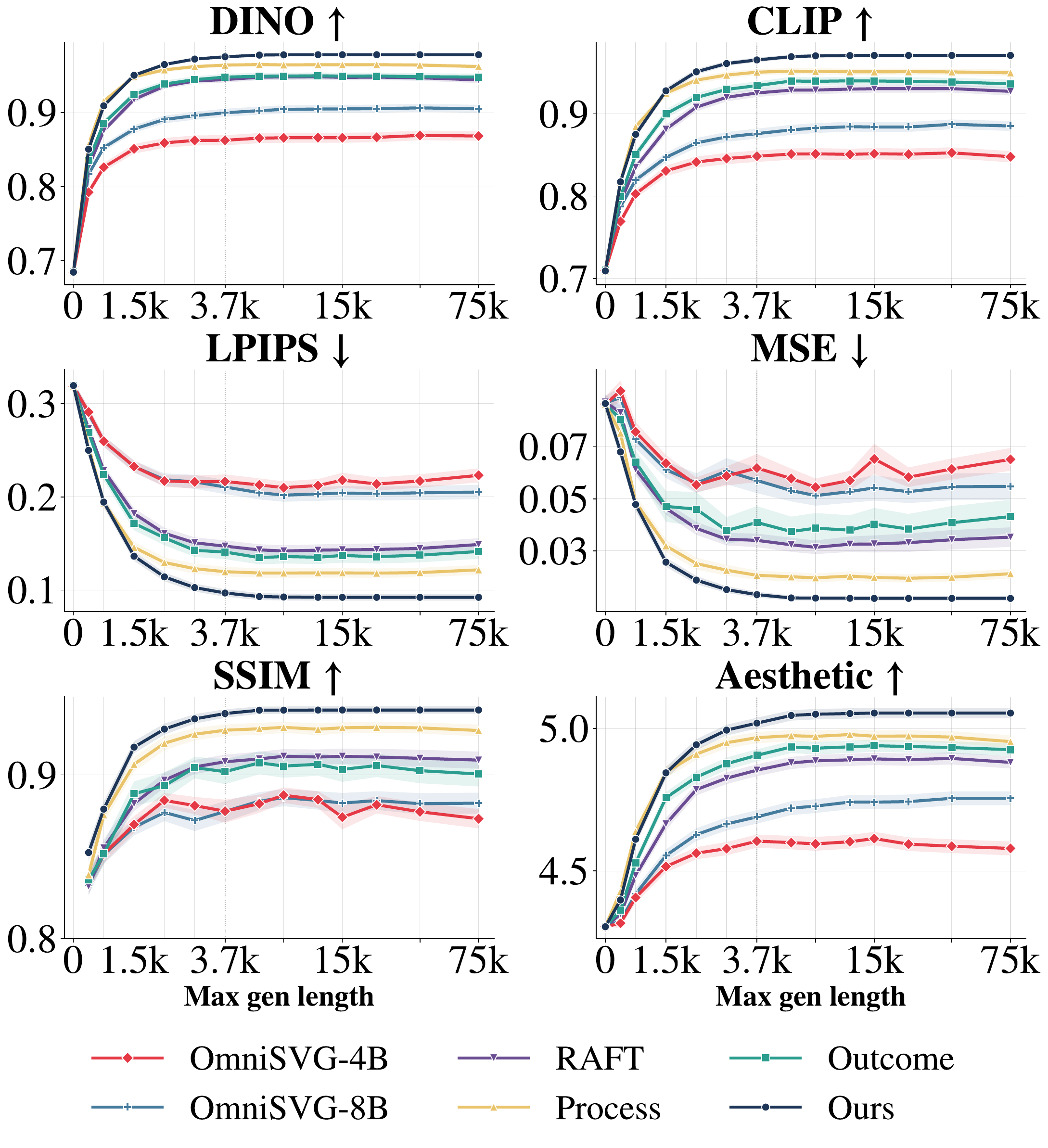}
\captionof{figure}{\textbf{Inference budget on MMSVGBench.}
Reconstruction quality across different Qwen token budgets. Performance saturates by 4k tokens, well below the 15k used in our main evaluation. Shaded areas represent S.E.M.}
\label{fig:tok_inference}
  \end{minipage}\hfill
  \begin{minipage}[t]{0.49\linewidth}
    \vspace{0pt}
    \centering
    \includegraphics[width=0.97\linewidth]{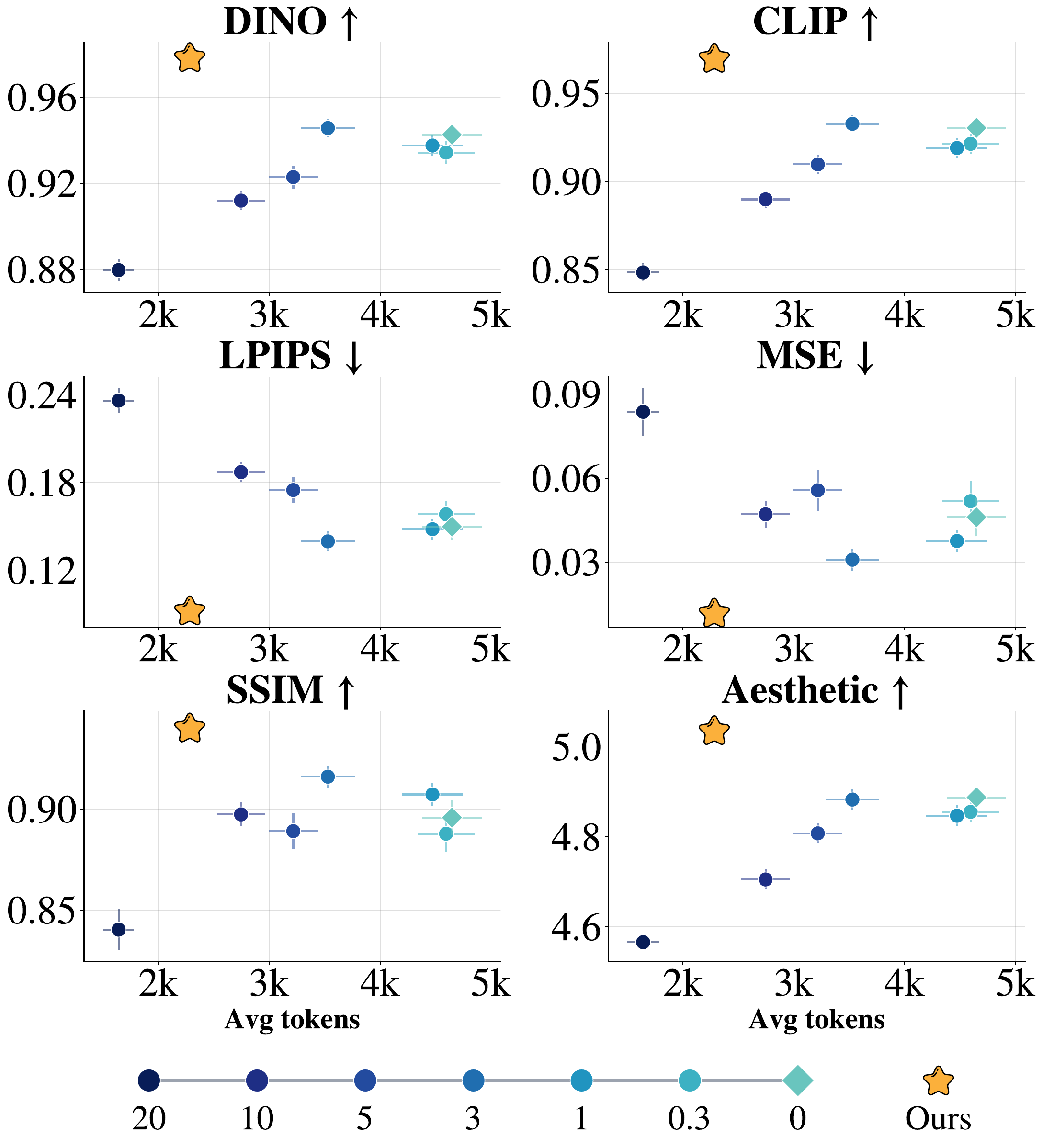}
\captionof{figure}{\textbf{Training with length penalties.} We augment outcome-only RL with a length penalty, and evaluate it on our MMSVGBench test set.
Point colors indicate the penalty weight, with the
visual reward weight fixed to one.
Zero denotes outcome-only RL without a length penalty.
Our method achieves better reconstruction across all tested weights, including settings producing shorter programs.
    }
    \label{fig:tok_train}
  \end{minipage}
\end{figure*}

Our method produces shorter programs, raising the question of how output length impacts reconstruction metrics. We examine whether baselines benefit from longer generation budgets or from explicit incentives for shorter programs.

\textbf{Inference budget.}
Our main evaluation (Tab.~\ref{tab:quant}) inference budget was 2k OmniSVG tokens, corresponding to roughly 15K Qwen tokens.
We extend the inference budget up to
10k OmniSVG tokens ($\approx$75k Qwen tokens).
Figure~\ref{fig:tok_inference} shows that our method outperforms all baselines across all budgets on our MMSVGBench test set.
Across methods, performance largely saturates by
approximately 4k Qwen tokens, below our standard
15k-token limit.
Nevertheless, methods' average token length differs dramatically, even after saturation. Specifically, we observe that the base model often generates repetitive
commands that increase program length without improving
reconstruction. 

Our outcome-only RL baseline provides no explicit incentive for shorter outputs, while our discounted process supervision favors earlier visual progress
(Sec.~\ref{sec:analysis}), which can encourage more
compact programs.
We therefore examine whether adding an explicit length
penalty can improve outcome-only RL. 

\textbf{Length penalties during training.}
Following RLRF~\citep{rodriguez2026rendering_rlrf},
we augment our outcome-only baseline with a length reward
that penalizes excessive length relative to the
ground-truth SVG:
\begin{equation}
R_{\text{len}} = 1 - \left( \frac{1}{L_{\text{gt}}} \max\left(0,\; L_{\text{pred}} - \frac{L_{\text{gt}}}{2}\right) \right)^{2},
\end{equation}
where $L_{\text{pred}}$ and $L_{\mathrm{gt}}$ denote the generated
and ground-truth SVG token lengths.
The combined reward is
\begin{equation}
R_{\mathrm{combined}}
=
R_{\mathrm{visual}} + w_{\mathrm{len}} R_{\mathrm{len}},
\end{equation}
where $R_{\mathrm{visual}}$ is the final-render reward
used by our outcome RL baseline. 
We fix its weight to one and vary
$w_{\mathrm{len}} \in \{0, 0.3, 1, 3, 5, 10, 20\}$,
with zero recovering the outcome RL baseline. 
Notably, this length reward requires additional access to ground truth code, unlike our method which uses only target images.

Figure~\ref{fig:tok_train} shows that a moderate penalty reduces average output length while improving several
reconstruction metrics on MMSVGBench.
Stronger penalties further shorten outputs but significantly degrade
reconstruction quality.
Our method achieves better quality across all six metrics
than every tested penalty setting, including the setting
that produces shorter programs than ours.
Thus, explicit length regularization can improve
outcome-only RL, but our method strikes a better quality--length tradeoff.

\section{Image-to-TikZ appendix}
\label{sec:setup_tikz_sm}

\subsection{Extended reward ablation}
\label{sec:tikz_ablation}
We extend the numeric evaluations in Tab.~\ref{tab:quant_tikz} with a reward composition ablation in Table~\ref{tab:ablation_tikz}, which demonstrates the same trend as the one observed in Image-to-SVG: process reward outperforms outcomde only, and the combination works best.

\setlength{\tabcolsep}{2pt}
  \begin{table}[t]
      \centering
      \resizebox{\linewidth}{!}{
      \begin{tabular}{lccccc}
          \toprule
          & DreamSim $\uparrow$ & SigLIP $\uparrow$ & CLIP $\uparrow$ & LPIPS $\downarrow$ & Tokens $\downarrow$ \\
          \midrule
          w/ Outcome                 & 80.9 & 90.6 & 89.3 & 37.8 & 1.3k \\
          w/ Process                 & \underline{81.1} & \underline{91.3} & \underline{90.2} & \underline{37.3} & \textbf{0.7k} \\
          w/ Both (Ours)             & \textbf{83.8} & \textbf{92.5} & \textbf{91.2} & \textbf{35.4} & \underline{1.0k} \\
          \bottomrule
      \end{tabular}
      }
      \caption{\textbf{Reward composition ablation on Image-to-TikZ.} Results at 30\% of the training schedule.}
      \label{tab:ablation_tikz}
  \end{table}

\subsection{Rendering boundaries}
\label{sec:tikz_closure}

After each generated token, we check whether it completes a TikZ
drawing statement. Commands such as \verb|\draw| and \verb|\fill|
end with a semicolon, so their boundaries are inexpensive to identify
in the generated text. In our default setting ($N=1$), we render the
prefix at every such boundary $b_j$. Roughly 5\% of generated tokens
mark these boundaries. Tokens between boundaries receive zero progress reward.

At a boundary, the latest drawing statement is complete, but the
prefix $y_{1:b_j}$ may still contain open braces, brackets, or
environments. The closure operator $\mathcal{C}$ makes the prefix
compilable by appending the corresponding closing delimiters and
environment endings in reverse nesting order. In
Fig.~\ref{fig:closure}b, for example, it appends \verb|\end{scope}|,
then \verb|\end{tikzpicture}|, then \verb|\end{document}|. We compile
the resulting program to render the figure as drawn so far. The
change in visual score since the preceding render,
$\Delta_j = F_j - F_{j-1}$, gives the progress reward at $b_j$.

\subsection{Training parameters}
\label{sec:tikz_train_params}
The visual similarity function $S$ between TikZ renders and a target is defined as Self-Sim, following prior work \citep{belouadi2024detikzify}. Self-Sim uses
DeTikZify-v2's own fine-tuned SigLIP vision encoder to extract patch-level features $f_{\text{in}}, f_{\text{pred}}$ for the target and rendered image, and
  computes their Earth Mover's Distance
  $d_{\text{EMD}}$ under patch-wise cosine cost:
  \begin{equation}
  S = 2\tanh\left(-d_{\text{EMD}}(f_{\text{in}}, f_{\text{pred}})\right) + 1,
  \label{eq:img_reward_tikz}
  \end{equation}
  $S$ lies within $[-1, 1]$, and for blank or malformed renders it is equal to $-1$. This score has been shown to correlate with human judgment of scientific
  figure reconstruction.

We fine-tune DeTikZify-8B with AdamW~\citep{loshchilov2019decoupled} and a constant learning rate scheduler, using a LoRA adapter~\citep{hu2022lora} ($r{=}64$, $\alpha_{\text{LoRA}}{=}128$, dropout $0.05$) on the attention (query, key, value, output) and MLP (gate, up, down) projections of the LLM decoder, leaving the vision encoder frozen. Each optimizer step uses 4 unique images with a group size of 32, resulting in 128 rollouts sampled at temperature $1.0$. 
We explore the learning rates $\text{lr} \in \{1\text{e-}5, 1\text{e-}4\}$ and the KL coefficients $\beta \in \{0, 0.01\}$, selecting $(\text{lr}, \beta)=(1\text{e-}5, 0)$, as also used by DeTikZify-v2.5 \cite{detikzify_v2_5_8b} to tune the same base model with GRPO.
Method-specific hyperparameters ($\alpha$, $\lambda$, $N$) are ablated one at a time using this configuration (Fig.~\ref{fig:analysis_tikz}). Each run is trained for 900 steps on a single NVIDIA B200 GPU. The training set consists of 22.4k samples from the DaTikZ-v3 train split~\citep{belouadi2024detikzify}.

\subsection{Baseline parameters}
\label{sec:tikz_baselines}

General-purpose (M)LLMs are prompted with the input image via each provider's API using the system prompt in Fig.~\ref{fig:sys_prompts}b. VinciCoder~\citep{zhao2025vincicoder} and
DeTikZify-v2~\citep{belouadi2024detikzify} are decoded with each model's default sampling parameters, with a generation budget of $4096$ tokens for all non-API models. All the post-training variants are reported with greedy sampling. RAFT~\citep{dong2023raft} is trained for 40 iterations, sampling 32 completions per input on a subset of 700
trainset images per iteration due to the high sampling runtime cost, selecting the highest-reward sample, and fine-tuning for two SFT epochs on the selected set.
  
\subsection{Metrics}
\label{sec:tikz_metrics}
SigLIP and CLIP use their large checkpoints (so400m/384, ViT-L/14). 
For fair comparison, token counts for all models use DeTikZify-v2's tokenizer, applied to the extracted TikZ code, and in the case of the API responses, the leaked prose and thinking-trace text, following the same convention as the SVG setup above.

\end{document}